\documentclass[letterpaper,journal]{IEEEtran}
\usepackage{amsmath,amsfonts}
\usepackage{array}
\usepackage[caption=false,font=normalsize,labelfont=sf,textfont=sf]{subfig}
\usepackage{textcomp}
\usepackage{stfloats}
\usepackage{url}
\usepackage{verbatim}
\usepackage{graphicx}
\usepackage[square,numbers]{natbib}
\usepackage{balance}

\makeatletter
\DeclareRobustCommand{\splitcite}[1]{%
  \begingroup
  \def\@first{}%
  \@for\@k:=#1\do{%
    \ifx\@first\@empty
      [\citenum{\@k}]%
      \gdef\@first{*}%
    \else
      , [\citenum{\@k}]%
    \fi
  }%
  \endgroup
}
\makeatother
\usepackage[ruled,vlined,linesnumbered]{algorithm2e} 
\usepackage{amsmath}
\usepackage{amssymb}
\usepackage{enumitem}
\usepackage{tabularx}
\usepackage{booktabs}
\usepackage{xcolor}
\usepackage{multirow}
\usepackage{pifont}

\usepackage{amsmath,amsthm}
\usepackage{thmtools,thm-restate}
\usepackage[colorlinks]{hyperref}

\theoremstyle{plain}
\declaretheorem[name=Theorem]{theorem}
\declaretheorem[name=Assumption]{assumption}

\hypersetup{
  colorlinks=true,  
  linkcolor=blue,   
  citecolor=blue,   
  urlcolor=blue,    
  filecolor=blue    
}

\begin{document}

\title{Task-agnostic Low-rank Residual Adaptation for Efficient Federated Continual Fine-Tuning}

\author{Feng Yu, Jia Hu,~\IEEEmembership{Member,~IEEE,} Geyong Min,~\IEEEmembership{Member,~IEEE}
\thanks{The authors are with the Department of Computer Science, University of Exeter, EX4 4PY Exeter, U.K. (e-mail: fy274@exeter.ac.uk; j.hu@exeter.ac.uk; g.min@exeter.ac.uk).}
}

\markboth{}%
{Shell \MakeLowercase{\textit{et al.}}: A Sample Article Using IEEEtran.cls for IEEE Journals}


\maketitle

\begin{abstract}
Federated Parameter-Efficient Fine-Tuning (Fed-PEFT) enables lightweight adaptation of large pre-trained models in federated learning settings by updating only a small subset of parameters. However, Fed-PEFT methods typically assume a fixed label space and static downstream tasks, which is restrictive in realistic application scenarios where clients continuously encounter new classes over time. This leads to an emerging problem, known as \emph{Federated Continual Fine-Tuning} (FCFT). 
In FCFT, clients collaboratively fine-tune a pre-trained model over a sequence of tasks, where each client observes disjoint sets of new classes over time, and task identity is unavailable at inference time.
FCFT is challenging because it simultaneously suffers from severe forgetting under non-IID client data distributions, parameter growth and task-specific inference caused by task-wise modules, and aggregation inconsistency across heterogeneous clients.
To address these challenges, we propose Federated Task-agnostic Low-rank Residual Adaptation (Fed-TaLoRA), a novel approach for efficient FCFT built on task-agnostic adaptation, post-aggregation model calibration, and strategic low-rank adaptation placement. Fed-TaLoRA continuously fine-tunes a single shared module across sequential tasks to avoid task-wise parameter growth, and further introduces a theoretically grounded residual weight update mechanism to calibrate the aggregated global model and improve aggregation fidelity. We provide a theoretical analysis of the convergence and aggregation behavior of Fed-TaLoRA. Extensive experiments on four benchmark datasets demonstrate that Fed-TaLoRA consistently outperforms strong baselines while reducing communication and computation costs significantly.

\end{abstract}


\section{Introduction}

\IEEEPARstart{F}{ederated} learning (FL) enables multiple clients to collaboratively train or adapt a shared model without centralizing raw data, making it attractive for privacy-sensitive and distributed applications. 
Recently, large pre-trained models \splitcite{han2021pre}, such as Vision Transformers (ViTs) \splitcite{dosovitskiy2020image} and BERT \splitcite{devlin2018bert}, have become widely adopted backbones for downstream tasks due to their strong transferability. However, fully fine-tuning such models in federated environments is often prohibitively expensive in communication, local computation, and storage. 
Federated Parameter-Efficient Fine-Tuning (Fed-PEFT) \splitcite{zhang2023fedpetuning,zhang2024towards} has therefore emerged as a practical paradigm for adapting pre-trained models in FL by updating only a small subset of parameters. 
Among representative PEFT methods, Low-Rank Adaptation (LoRA) \splitcite{hu2022lora} freezes the pre-trained weights and inserts trainable low-rank matrices into model layers, making it particularly attractive for communication- and resource-constrained federated environments. 
Therefore, Fed-PEFT substantially reduces communication, computation, and storage costs while preserving the benefits of pre-training \splitcite{babakniya2023sloraa}.

However, existing Fed-PEFT methods still focus on a fixed label space and static downstream tasks, implicitly assuming that the label set observed by clients does not expand over time. This assumption is restrictive in realistic application scenarios, where clients may continuously encounter previously unseen categories and still need to preserve previously learned knowledge \splitcite{wang2024comprehensive,gong2024ode,yue2025ug}. 
Class-Incremental Learning (CIL) tackles this issue by learning new classes while retaining competence on previously learned ones \splitcite{zhou2024class}. When extended to federated settings, Federated Class-Incremental Learning (FCIL) requires the global model to continually incorporate newly arriving classes while coping with non-independently and identically distributed (IID) client distributions and communication constraints \splitcite{dong2022federateda,wang2024federated}. Such a capability is particularly relevant to Edge-AI applications, where distributed devices operate in dynamic environments, encounter new classes over time, and must adapt without raw-data sharing or frequent centralized retraining. 
In this paper, we study this problem from the perspective of adapting pre-trained models and term it \emph{Federated Continual Fine-Tuning (FCFT)}. 
In FCFT, a pre-trained model is collaboratively fine-tuned over a sequence of incremental tasks. At each task, clients observe samples from a new disjoint set of classes. During inference, the model must distinguish among all classes observed so far without access to task identity.

Despite recent progress, an effective solution for FCFT is still lacking because of the following challenges. 
First, FCFT aggravates catastrophic forgetting of previously learned classes. Since each client only observes newly arriving classes and cannot revisit old data, local updates become biased toward the current classes, which causes classifier bias and degrades performance on old classes. 
This issue is further amplified in federated settings, where non-IID client distributions push local updates toward client-specific optima rather than the global objective, leading to client drift \splitcite{sun2024improvinga}.
Second, many existing FCFT methods encode knowledge in task-specific modules, such as prompt pools, adapter pools, or per-task LoRA parameters. As the number of tasks increases, these designs incur parameter growth and often require task-specific module selection during inference \splitcite{bagwe2023fedcprompta,halbe2024continual,liu2023fedetd,guo2024pilora}.
Third, LoRA-based federation introduces an additional aggregation challenge. Each client jointly optimizes two coupled low-rank factors, whereas the server typically averages them factor-wise. 
Since the product of averaged factors is generally not equal to the average of client-wise products, the aggregated LoRA update deviates from the client-averaged dense update that FedAvg aims to approximate \splitcite{mcmahan2017communication}. 
Under heterogeneous clients, this mismatch can accumulate over rounds and further worsen client drift \splitcite{sun2024improvinga,wang2024flora}.

To address these challenges, we propose \textbf{Fed}erated \textbf{T}ask-\textbf{a}gnostic \textbf{Lo}w-rank \textbf{R}esidual \textbf{A}daptation (Fed-TaLoRA), a novel approach for efficient federated continual fine-tuning. 
Fed-TaLoRA continuously updates a single shared LoRA module across sequential tasks, thereby avoiding task-wise parameter growth. 
To improve aggregation fidelity under heterogeneous clients, it further introduces a residual weight update mechanism for post-aggregation calibration. 
In addition, separate learning rates are adopted for the representation and classification layers to balance adaptation stability and efficiency.

Our main contributions are as follows:
\begin{itemize}
    \item We propose Fed-TaLoRA, which tunes only task-agnostic LoRA parameters in federated continual fine-tuning with a theoretically grounded residual weight update mechanism.
    Fed-TaLoRA ensures accurate model aggregation to mitigate client drift in heterogeneous data settings while eliminating the need for keeping task-specific parameters, thereby delivering superior performance and efficiency in resource-constrained FCFT scenarios.
    \item We provide a theoretical analysis of Fed-TaLoRA, including its convergence behavior and LoRA aggregation properties. In particular, we show that the proposed residual weight update recovers the corresponding client-averaged dense LoRA update under our aggregation formulation, thereby improving aggregation fidelity in heterogeneous federated settings.
    \item The proposed Fed-TaLoRA method has three novel mechanisms: 
    (1) \textit{task-agnostic adaptation} that enables all tasks and clients to collaboratively update a single shared LoRA module to avoid parameter redundancy and improve generalization;
    (2) \textit{post-aggregation model calibration} that employs a residual weight update mechanism to correct aggregation bias and reconstruct the global model's weight trajectory under non-IID settings;
    and (3) \textit{strategic LoRA placement} that inserts LoRA modules into attention and feedforward layers of early transformer blocks to balance adaptation capacity with communication and computational efficiency.
    \item We evaluate the performance of our method on four popular benchmark datasets. 
    Extensive experiments demonstrate that Fed-TaLoRA outperforms state-of-the-art methods while reducing communication and computation costs by up to 53\% and 14\%, respectively. 
\end{itemize}

The rest of this paper is organized as follows:
Section \ref{related_work} reviews the related work including federated class-incremental learning (FCIL), Fed-PEFT and FCFT.
Section \ref{sec:preliminaries} gives the formal definition of FCFT.
The proposed Fed-TaLoRA is introduced in Section \ref{sec:methods}.
The rigorous theoretical analysis for our approach is given in Section \ref{sec:theory_analysis}.
Section \ref{sec:experiments} and Section \ref{sec:fur_analysis} present the simulation results and further performance analysis.
Finally, Section \ref{sec:conclusions} concludes the paper.

\section{Related Work}
\label{related_work}

In this section, we review the related work from three perspectives: Fed-PEFT, FCIL and FCFT.

\textbf{Fed-PEFT} adapts PEFT techniques to federated environments with resource constraints \splitcite{zhang2023fedpetuning,babakniya2023sloraa,chen2022fedtune}. 
Research revealed that directly aggregating local LoRA parameters introduces noise in non-IID settings \splitcite{babakniya2023sloraa,sun2024improvinga}.
Proposed solutions either constrain the model capacity by freezing initialized matrices \splitcite{sun2024improvinga,hao2024flora} or increase communication costs through stacking-based aggregation \splitcite{wang2024flora}. 
Through the residual weight update, our approach achieves accurate parameter aggregation without substantial overhead while effectively addressing challenges in incremental tasks. 
Its effectiveness is similar to the recent FedEx-LoRA \splitcite{singhal2025fedex}, with the difference that we focus on ViT for FCFT rather than language models within traditional FL.

\textbf{FCIL} enables privacy-preserving model adaptation to new classes across distributed clients. 
Early approaches like GLFC \splitcite{dong2022federateda} and LGA \splitcite{dong2024no} relied on privacy-compromising rehearsal buffers \splitcite{nori2025federated}.
Privacy-preserving alternatives emerged through generative models (TARGET \splitcite{zhang2023targeta}) and prototypical knowledge transfer (FedProK \splitcite{gao2024fedprok}), but introduced significant computational overhead.
Recent pre-trained model adaptations achieve better performance with reduced computational costs, yet face trade-offs between inference efficiency and memory requirements \splitcite{liu2023fedetd, bagwe2023fedcprompta,halbe2024continual,guo2024pilora,salami2024closed}.
More recently, 
LoRM \splitcite{salami2024closed} proposes a closed-form merging method for parameter-efficient modules.

\textbf{FCFT} approaches \splitcite{liu2023fedetd, bagwe2023fedcprompta, halbe2024continual} encode task-specific knowledge in modular parameters, reducing forgetting but introducing inference delays through similarity matching. 
PILoRA \splitcite{guo2024pilora} addresses latency issues but requires additional prototype parameters. 
Recently, LoRM \splitcite{salami2024closed} proposes a closed-form solution for merging LoRA for FCFT, while pFedMxF \splitcite{zhang2025pfedmxf} introduces orthogonal frequency component decomposition to address spatial, temporal, and resource heterogeneity. 
Existing approaches following the \textit{task-specific fine-tuning followed by adaptation} pattern may be suboptimal for FCFT settings with non-IID data \splitcite{liao2023parameterefficient,wu2025sd}. 
Inspired by SLCA \splitcite{zhang2024slca}, Fed-TaLoRA diverges from this paradigm by continuously fine-tuning task-agnostic LoRA parameters with a smaller learning rate for representation layers than classification layers, enabling effective knowledge transfer while balancing performance and resource efficiency.

\section{Preliminaries}
\label{sec:preliminaries}
\textbf{Federated Continual Fine-Tuning (FCFT)} is a challenging and realistic setting where a global model is collaboratively learned over 
a sequence of incremental tasks that introduce new classes across different clients.
In this setting, suppose that there are $K$ clients and a global server $\mathcal{S}$, each client $k \in [1, K]$ trains its local model parameters $\theta_k^t$ over its own stream dataset $\mathcal{D}_{k}=\{\mathcal{D}_k^t\}_{t=1}^T$, which consists of a sequence of $T$ incremental tasks $\{1, 2, ..., t, ..., T\}$.
For $t$-th task, the dataset of $k$-th client $\mathcal{D}_k^t = \{\mathcal{X}_k^t, \mathcal{Y}_k^t\}$ contains $N_k^t$ pairs of the training sample $x_i^t$ and corresponding label $y_i^t$, where $\mathcal{Y}_k^t \in \mathcal{C}_k^t$ and $\mathcal{C}$ denotes the class set.
Importantly, the class sets of different tasks are disjoint and the distribution of different clients within the same task is non-IID.
Suppose that $\mathcal{L}$ is a pre-defined loss function on the current dataset $\mathcal{D}_k^t$, the objective of local clients is to avoid interference and absorb knowledge from previously learned tasks, which can be formulated as follows:
\begin{equation}
    \arg \min\limits_{\theta_k^t} \mathcal{L}(\theta_k^t; \theta_s^{t-1}, \mathcal{X}_k^t, \mathcal{Y}_k^t),
\end{equation}
where $\theta_s^{t-1}$ is the global model of the previous task.
After the fine-tuning for the current task is complete, each client $k$ transmits the fine-tuned model parameters $\theta_k^t$ to the server $\mathcal{S}$, and server $\mathcal{S}$ aggregates them into the global model $\theta_s^{t+1}$ with a weighting parameter $\omega_k$ to integrate the task knowledge across all clients as follows: 
\begin{equation}
    \theta_s^{t+1} = \sum_{{k=1}}^K \omega_k \theta_k^t,\\\ \text{where} \\\ \omega_k = \frac{N_k^t}{\sum_{k=1}^K N_k^t}.   
    \label{eq:averaging}
\end{equation}
Then, the server $\mathcal{S}$ distributes the global parameter $\theta_s^{t+1}$ to all clients for learning the next task.
After learning all tasks, the final global model aims to classify test samples of all seen classes and address the data heterogeneity while maintaining a low communication cost.
Notably, in FCFT, the task identity is not known at inference.
For ease of expression, we list some important notations of this paper in Table \ref{tab:key-notations}.

\begin{figure*}[!ht]
    \centering
    \includegraphics[width=0.8\linewidth]{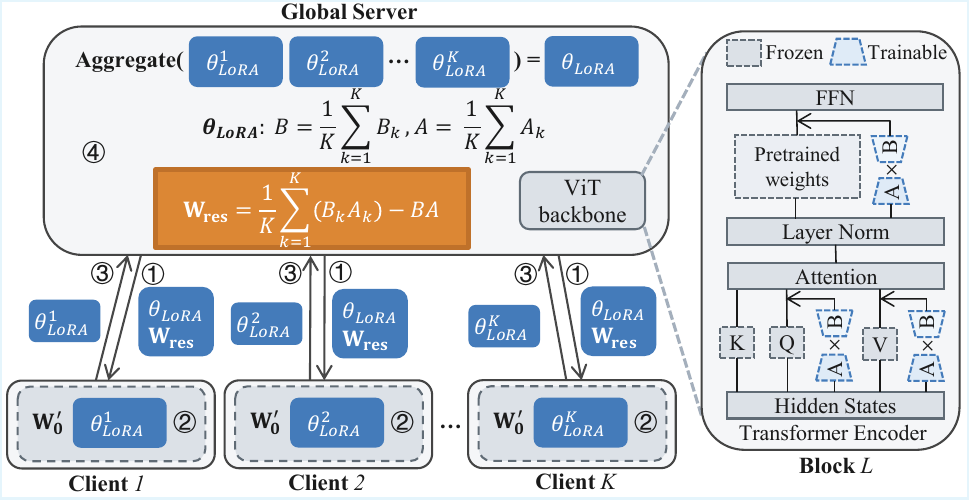}
    \caption{Pipeline of Fed-TaLoRA for FCFT.
    Clients first receive the global model and fine-tune only their local LoRA parameters embedded in attention layers and FFN (\ding{172} $\rightarrow$ \ding{173}) on their own private data containing new classes, 
    then upload these fine-tuned model parameters (\ding{174}) for server-side aggregation (\ding{175}). 
    Additionally, the server computes residual weights to capture cross-client variations, preparing an enhanced model of the next training round.
    }
    \label{fig:overview}
\end{figure*}

\begin{table}[t]
\renewcommand{\arraystretch}{1.15}
\caption{Key Notations}
\label{tab:key-notations}
\centering
\begin{tabular}{@{} l p{0.70\linewidth} @{}}
\toprule
\textbf{Notation} & \textbf{Semantics} \\
\midrule
$T$                  & length of incremental tasks \\
$R$                  & communication round in FCFT  \\
$K$                  & number of local clients  \\
$\varepsilon$        & local epochs on each client \\
$\theta_f$           & frozen representation layer's parameters \\
$\mathcal{D}_k^t$    & training dataset on client $k$ at task $t$\\
$\mathcal{C}$        & class set \\
$\mathcal{C}_k^t$    & class set owned by client $k$ at task $t$ \\
$\mathbf{W}_0$       & frozen pre-trained weights in PTM  \\
$\mathbf{W}_{res}$   & residual weights computed at global server $\mathcal{S}$ \\
$\theta_{LoRA}^{k,r}$    & LoRA updates on client $k$ at round $r$ \\
$\eta_1$               & learning rate for $\theta_{LoRA}^k$ \\
$\eta_2$               & learning rate for $\theta_{cls}$ \\
$\theta_{LoRA}^k$         & trainable LoRA parameters \\
$\theta_{cls}$       & classification layer's parameters \\
$\theta_{LoRA}$      & aggregated LoRA parameters \\
$\mathbf{B}, \mathbf{A}$ & two low-rank matrices constituting $\theta_{LoRA}$ \\ 
$\mathbf{W}_k$         & trained weights on client $k$ \\
$\mathbf{W}^{*}_g$     & optimal global model \\
$\hat{\mathbf{W}}_g$   & full model on the server \\

\bottomrule
\end{tabular}
\end{table}

\section{Our Approach}
\label{sec:methods}

In this section, we introduce proposed Fed-TaLoRA composed of task-agnostic LoRA and residual weight of update of pre-trained models. 

\subsection{Continually Fine-Tune Task-Agnostic LoRA of Pre-Trained Models}
\textbf{Task-Agnostic LoRA}.
Inspired by \splitcite{hu2021loraa,zhang2024slca}, we incorporate task-agnostic LoRA (TaLoRA) of PTMs into the federated continual fine-tuning paradigm.
Our approach is based on the hypothesis that weight updates during adaptation possess a low ``intrinsic rank''.
We denote the pre-trained weight matrix $\mathbf{W} \in \mathbb{R}^{d \times k}$ and the parameter update shared across all tasks and clients as $\Delta \mathbf{W}$.
To reduce computational complexity, we constrain the parameter updates by representing $\Delta \mathbf{W}$ through a low-rank decomposition during different task phases:
\begin{equation}
\mathbf{W} + \Delta \mathbf{W} = \mathbf{W} + \mathbf{B}\mathbf{A}.
\label{eq:vanilla-lora}
\end{equation}
Here, $\mathbf{B} \in \mathbb{R}^{d \times r}$ and $\mathbf{A} \in \mathbb{R}^{r \times k}$, where rank $r \ll \min(d, k)$. 
We initialize matrix $\mathbf{B}$ with zeros and matrix $\mathbf{A}$ with random Gaussian values, ensuring that $\Delta \mathbf{W} = \mathbf{B} \mathbf{A}$ equals zero at the beginning of training at the server side.
The pre-trained ViT backbone \splitcite{dosovitskiy2020image} with these embedded low-rank matrices $\mathbf{B}$ and $\mathbf{A}$ is then distributed to all participating clients for local training.
During local training, only the two low-rank matrices $\mathbf{B}$ and $\mathbf{A}$ can be trained, rather than the complete set of weight parameters in attention layers and feed-forward networks (FFN).
This approach restricts $\Delta \mathbf{W}$ to rank $r$ and reduces the total number of trainable parameters from $d \cdot k$ to $r \cdot (d + k)$, thereby avoiding the computational expense of fully fine-tuning the PTM weights.
Importantly, all clients collaborate on training shared task-agnostic LoRA parameters across incremental tasks, rather than maintaining separate LoRA parameters for each task in each communication round.

We argue that the weight parameters in the attention layers and the FFN modules incur substantial resource usage, including computational and communication costs in FL environments.
Our empirical investigations in Section \ref{appendix:lora_mlp} reveal that extending LoRA to the fully connected layers within FFNs substantially improves model performance, aligning with findings reported by \splitcite{fomenko2024note,lin2024tracking,agiza2024mtlora}.
Consequently, our approach enables all participating clients to continuously fine-tune task-agnostic LoRA parameters embedded in both attention layers and fully connected layers throughout the incremental learning process.

\subsection{Residual Weight Update of Pre-Trained Models}
When deploying LoRA-based methods in federated settings with non-IID data, naive parameter aggregation introduces mathematical inconsistencies that degrade model performance \splitcite{babakniya2023sloraa,sun2024improvinga}.
This occurs because independently averaging low-rank matrices $\mathbf{A}$ and $\mathbf{B}$ across clients creates a critical discrepancy.

To achieve accurate aggregation with low additional overhead, we introduce the Residual Weight Update (ResWU) mechanism.
Using the same aggregation weights as in Eq.~\eqref{eq:averaging}, the server first aggregates the low-rank factors as
\[
\mathbf{B}=\sum_{k=1}^{K}\omega_k\mathbf{B}_k,
\qquad
\mathbf{A}=\sum_{k=1}^{K}\omega_k\mathbf{A}_k,
\]
where $\sum_{k=1}^{K}\omega_k=1$.
The corresponding residual weight is then defined as
\begin{align}
\label{eq:residual_weights}
\mathbf{W}_{res}
&=
\sum_{k=1}^{K}\omega_k(\mathbf{B}_k\mathbf{A}_k)
-
\Big(\sum_{k=1}^{K}\omega_k\mathbf{B}_k\Big)
\Big(\sum_{k=1}^{K}\omega_k\mathbf{A}_k\Big) \\
&=
\sum_{k=1}^{K}\omega_k(\mathbf{B}_k\mathbf{A}_k)-\mathbf{B}\mathbf{A}.
\label{eq:res}
\end{align}
This residual weight is distributed to clients together with the aggregated LoRA parameters.
Clients then update their frozen pre-trained weights by
\begin{equation}
\label{eq:ResWU}
\mathbf{W}_0'=\mathbf{W}_0+\mathbf{W}_{res},
\end{equation}
where $\mathbf{W}_0$ denotes the original pre-trained weights.
The effective global dense model after residual correction can be written as
\begin{align}
\label{eq:w_g}
\mathbf{W}_g
=
\mathbf{W}_0'+\mathbf{B}\mathbf{A}
=
\mathbf{W}_0+\mathbf{W}_{res}+\mathbf{B}\mathbf{A}.
\end{align}
Substituting Eq.~\eqref{eq:res} into Eq.~\eqref{eq:w_g} yields
\begin{align}
\label{eq:w_g*}
\mathbf{W}_g^*
=
\mathbf{W}_0+\sum_{k=1}^{K}\omega_k(\mathbf{B}_k\mathbf{A}_k).
\end{align}
Therefore, ResWU exactly recovers the weighted dense aggregation target under the same FedAvg-style weights used by the server.
Note that $\mathbf{W}_{res}=0$ only when the client adapters are identical (or sufficiently close), rather than merely when the client data are IID.
This formulation demonstrates that Fed-TaLoRA, through ResWU, faithfully reconstructs the average global weight trajectory even under non-IID client updates, addressing the critical inconsistency in naive LoRA aggregation.
\begin{algorithm}[!htp]\small
    \caption{Training Algorithm of Fed-TaLoRA}
    \label{alg:algo_Fed-TaLoRA}
    \KwIn{Training dataset $\mathcal{D}_k^t$ for each client $k$ at task $t=1,...,T$; communication round $R$; local epoch $\varepsilon$; global model $M_\theta(\cdot)=h_{\theta_{cls}}(f_{\theta_{rps}}(\cdot))$ with $\theta_{rps}=\{\theta_f, \theta_{LoRA}\}$; learning rates $\eta_1$ for $\theta_{LoRA}$ and $\eta_2$ for $\theta_{cls}$ ($\eta_1 < \eta_2$)}
    \KwOut{Final model $M_\theta^*$}

    Initialize $\theta_{rps}$ from $W_0$, randomly initialize $\theta_{cls}$ \;
    Initialize global model $\theta^0 = \{\theta_{rps}, \theta_{cls}\}$ and send to all clients \;
    
    \For{each task $t = 1, \ldots, T$}{
        \For{round $r = 1, \ldots, R$}{
            $U^r \gets U$ \tcp*[f]{full client participation} \;
            \For{each client $k \in U^r$ in parallel}{
                Update $\mathbf{W}_0$ with $\mathbf{W}_{res}$ by Eq. (\ref{eq:ResWU}) \;
                Assemble $\theta_k^r \gets \{\theta_f, \theta_{LoRA}^r\}$ \;
                \For{epoch $e = 1$ to $\varepsilon$}{
                    Update $\theta_{LoRA}^{k,r}$ and $\theta_{cls}$ on $\mathcal{D}_k^t$ with learning rates $\eta_1$, $\eta_2$ \;
                }
                Send $\theta_{LoRA}^{k,r}$ to the server \;
            }
            /* For global server */\\
            Receive all $\{\theta_{LoRA}^{k,r}\}_{k \in U^r}$ \;
            Aggregate $\theta_{LoRA}^{r}$ for subsequent learning \;
            Compute residual weights $\mathbf{W}_{res}$ using Eq.~(\ref{eq:residual_weights}) \;
        }
    }

    \Return $M^*_{\theta}$
\end{algorithm}

Our \textbf{Fed-TaLoRA} learning procedure follows a systematic communication protocol between the server and clients. The process begins with the server initializing a pre-trained model with trainable task-agnostic LoRA parameters, $\theta_{LoRA}$ and frozen backbone parameters $\theta_f$, which are then distributed to all participating clients. Upon receiving the model, each client first applies the computed residual weight $\mathbf{W}_{res}$ to update the pre-trained weights $\mathbf{W}_0$, thereby ensuring mathematical consistency across the federation. Clients then fine-tune the task-agnostic LoRA parameters on their private datasets through a sequence of incremental tasks. After local training, only the updated LoRA parameters are sent back to the server, where they are aggregated via weighted averaging to consolidate learned knowledge. The server then computes a new residual weight for the subsequent communication round. This iterative process continues until model convergence. Notably, in IID settings, $\mathbf{W}_{res}$ simplifies to zero. Figure~\ref{fig:overview} illustrates the complete Fed-TaLoRA pipeline, with the formal algorithm provided in Algorithm~\ref{alg:algo_Fed-TaLoRA}.

\section{Convergence Analysis of Fed-TaLoRA}
\label{sec:theory_analysis}
In this section, we analyze Fed-TaLoRA under full client participation on the current task.
For theoretical consistency with server update, $\theta^r$ denotes the global LoRA parameter maintained and aggregated by the server at the communication round $r$.
The locally updated classifier is omitted from $\theta$ because it is not aggregated by the server.
Let $\mathcal{L}_k(\theta)$ denote the empirical loss of client $k$ on the current task, and let
$\mathcal{L}(\theta)=\sum_{k=1}^{K}\omega_k\mathcal{L}_k(\theta)$
be the global objective, where $\omega_k$ is defined in Eq.~\eqref{eq:averaging}.
Throughout the analysis we adopt the standard assumptions used in nonconvex local-SGD/FedAvg analyses \splitcite{li2020federated,khaled2020tighter,fallah2020personalized}.

\begin{assumption}[Smoothness]
\label{ass:smooth_full}
Each $\mathcal{L}_k$ is differentiable and $L$-smooth:
$\|\nabla\mathcal{L}_k(\theta_1)-\nabla\mathcal{L}_k(\theta_2)\|\le L\|\theta_1-\theta_2\|$ for all $\theta_1,\theta_2$.
\end{assumption}

\begin{assumption}[Unbiased Gradients with Bounded Variance]
\label{ass:variance_full}
For stochastic gradients $g_k(\theta,\xi)$ evaluated on mini-batch $\xi$,
$\mathbb{E}_\xi[g_k(\theta,\xi)]=\nabla\mathcal{L}_k(\theta)$ and
$\mathbb{E}_\xi\|g_k(\theta,\xi)-\nabla\mathcal{L}_k(\theta)\|^2\le\sigma^2$.
\end{assumption}

\begin{assumption}[Bounded Client Drift]
\label{ass:hetero_full}
There exists $\delta\ge0$ such that
$\|\nabla\mathcal{L}_k(\theta)-\nabla\mathcal{L}(\theta)\|\le\delta$ for all $k,\theta$.
\end{assumption}

Let $\theta^{r}$ be the global LoRA iterate kept by the server at communication round $r$, and let $\theta_{k}^{r,e}$ denote the $e$-th local LoRA iterate on client $k$ ($e=0,\dots,\varepsilon$), with $\theta_{k}^{r,0}=\theta^{r}$. Clients update their local LoRA parameters with step size $\eta$.

\begin{restatable}[One-Round Progress]{lemma}{MainLemma}
\label{lem:one_round_full}
Under Assumptions~\ref{ass:smooth_full}--\ref{ass:hetero_full}, choose $\eta\le\frac{1}{4L\varepsilon}$. For any round $r$,
\begin{align*}
    \mathbb{E}\big[\mathcal{L}(\theta^{r+1})\mid\theta^{r}\big]
    &\le \mathcal{L}(\theta^{r})
      -\frac{\eta\varepsilon}{2}\big\|\nabla\mathcal{L}(\theta^{r})\big\|^2 
      + \eta^2 L \varepsilon\big(\sigma^2 + \varepsilon\delta^2\big),
\end{align*}
where the expectation is taken over mini-batch randomness.
\end{restatable}

\begin{theorem}[Stationarity Rate]
\label{thm:stationarity_full}
Under Assumptions~\ref{ass:smooth_full}--\ref{ass:hetero_full} and $\eta\le\frac{1}{4L\varepsilon}$, running Fed-TaLoRA for $R$ rounds yields
\[
\frac{1}{R}\sum_{r=0}^{R-1}
\mathbb{E}\big[\|\nabla\mathcal{L}(\theta^{r})\|^2\big]
\le 
\frac{2(\mathcal{L}(\theta^{0})-\mathcal{L}^\star)}{\eta\varepsilon R}
+ 2\eta L(\sigma^2+\varepsilon\delta^2),
\]
where $\mathcal{L}^\star$ is the infimum of $\mathcal{L}$.
\end{theorem}

Please find the proof of Lemma~\ref{lem:one_round_full} and Theorem~\ref{thm:stationarity_full} in Appendix \ref{app:proof_convergence}.

Theorem~\ref{thm:stationarity_full} shows that Fed-TaLoRA attains an $\mathcal{O}(1/R)$ stationarity rate with a bias controlled by $\eta(\sigma^2+\varepsilon\delta^2)$, matching the behavior of other nonconvex federated learning methods under comparable assumptions \splitcite{li2020federated}. In our analysis, the role of ResWU is to remove the additional factor-wise LoRA aggregation bias so that the aggregated update follows the corresponding weighted dense aggregation target, which is consistent with the faster convergence observed empirically in Fig.~\ref{fig:convergence}.

\section{Experiments}
\label{sec:experiments}
In this section, we will introduce the experimental settings and the results, ablation studies of proposed Fed-TaLoRA and corresponding resource usage analysis.

\subsection{Experimental Settings}
\textbf{Benchmark.}
Following \splitcite{guo2024pilora}, we use widely adopted datasets in the field of federated class-incremental learning including 
\textit{(1) CIFAR-100} \splitcite{krizhevsky2009learninga}: This dataset comprises images from 100 classes. 
These classes are divided into randomly 10 separate incremental tasks, with each task featuring a distinct set of classes.
\textit{(2) Tiny-ImageNet} \splitcite{le2015tiny}: This dataset is composed of images from 200 classes. It includes challenging examples from the original ImageNet \splitcite{deng2009imagenet}. 
These classes are also randomly divided into 10 distinct incremental tasks.
\textit{(3) ImageNet-Subset} \splitcite{hu2021distilling}: This dataset contains the first 100 classes of the arranged 1,000 classes. 
For each task, we split the training data of the above three datasets into a training set and a validation set in an 8:2 ratio, and we evaluate the global model using a global test set.
\textit{(4) ImageNet} \splitcite{deng2009imagenet}: This dataset consists of 1,000 classes. For simplicity, we first preprocessed the dataset, i.e., keeping 700 images for each class. 
Then we split the training data of the dataset into a training set and a validation set in an 5:1 ratio, and evaluate the global model using a global test set.
For a fair comparison with the baseline class-incremental learning methods in the FCFT setting, we follow the same protocols proposed by \splitcite{zhou2023pycila} to split 10 incremental tasks in our main experiments, and utilize the same class order generated from \splitcite{zhou2023pycila}.

\label{appendix:non_iid}
\begin{figure}[!ht]
    \centering
    \subfloat{
    \centering
    \includegraphics[width=0.48\linewidth]{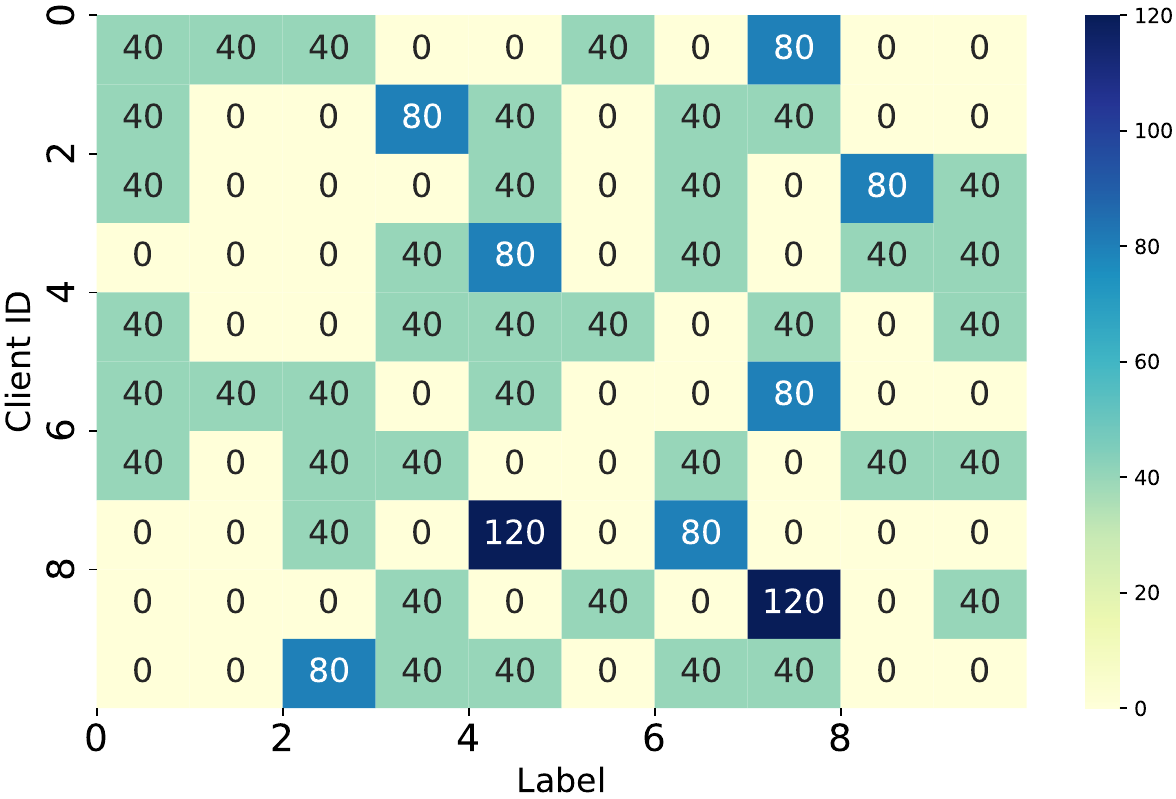}}
    \hfill
    \subfloat{
    \centering
    \includegraphics[width=0.48\linewidth]{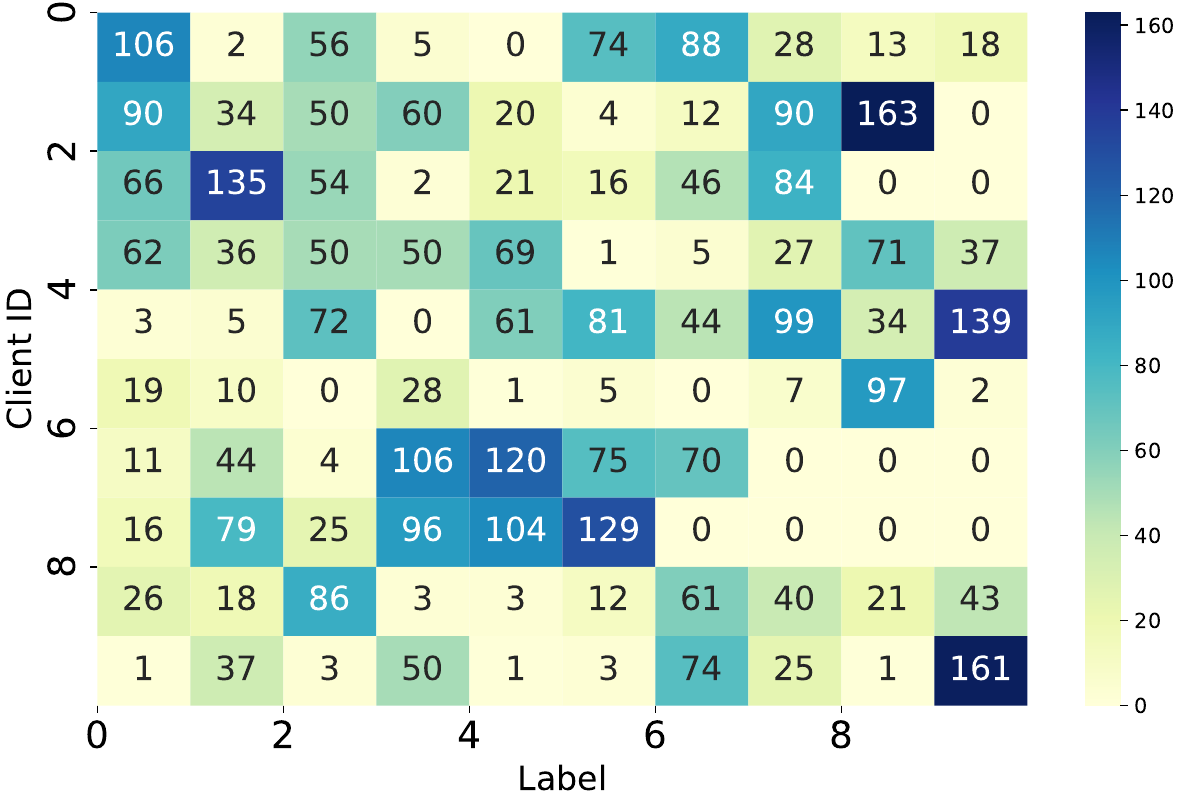}}
    \caption{An example of the non-IID setting on CIFAR-100 dataset. The value in each rectangle is the number of data samples of a class belonging to a certain client. (left: $\alpha=6$, right: $\beta=0.5$.)}
    \label{fig:non_iid}
\end{figure}

\begin{table*}[!ht]
\caption{
Results (\%) on CIFAR-100 including 10 tasks (10 classes per task) under two types of non-IID settings. $\alpha$ denotes the number of distinct labels assigned to each client, while $\beta$ represents the concentration parameter of the Dirichlet distribution. 
Results marked with $\dagger$ are from \splitcite{guo2024pilora}, and those marked with $\dagger\dagger$ are from \splitcite{zhang2025pfedmxf}.
}
\label{tab:cifar100}
\centering
\scalebox{1.2}{
\begin{tabular}{cc cc cc cc cc cc c} 
\hline
\multicolumn{1}{c}{\bf{Non-IID}} & \multicolumn{6}{c}{\bf{Quantity-based label imbalance}} & \multicolumn{6}{c}{\bf{Distribution-based label imbalance}} \\ \cline{2-13} 
\multicolumn{1}{c}{\bf{Partition}} & \multicolumn{2}{c}{$\alpha=6$} & \multicolumn{2}{c}{$\alpha=4$} & \multicolumn{2}{c}{$\alpha=2$} & \multicolumn{2}{c}{$\beta=0.5$} & \multicolumn{2}{c}{$\beta=0.1$} & \multicolumn{2}{c}{$\beta=0.05$} \\ \cline{2-13}
\bf{Method} & $FAA$ & \multicolumn{1}{c}{$AIA$} & $FAA$ & \multicolumn{1}{c}{$AIA$} & $FAA$ & $AIA$ & $FAA$ & \multicolumn{1}{c}{$AIA$} & $FAA$ & \multicolumn{1}{c}{$AIA$} & \multicolumn{1}{c}{$FAA$} & \multicolumn{1}{c}{$AIA$} \\ 
\hline
\multicolumn{1}{c}{Joint}  
& 88.6 & \multicolumn{1}{c}{-} & 84.3 & \multicolumn{1}{c}{-} & 79.8 & - & 90.1 & \multicolumn{1}{c}{-} & 87.8 & \multicolumn{1}{c}{-} & 85.9 & - \\
\multicolumn{1}{c}{EWC\textsuperscript{\textdagger}}  
& 57.9 & \multicolumn{1}{c}{69.1} & 55.9 & \multicolumn{1}{c}{66.8} & 42.2 & 52.7 & 65.5 & \multicolumn{1}{c}{77.8} & 57.8 & \multicolumn{1}{c}{73.2} & 43.5 & 59.2 \\
\multicolumn{1}{c}{LwF\textsuperscript{\textdagger}}  
& 57.4 & \multicolumn{1}{c}{68.8} & 55.1 & \multicolumn{1}{c}{66.7} & 40.8 & 52.9 & 64.7 & \multicolumn{1}{c}{77.5} & 54.6 & \multicolumn{1}{c}{63.3} & 45.7 & 64.5 \\
\multicolumn{1}{c}{iCaRL\textsuperscript{\textdagger}}  
& 35.8 & \multicolumn{1}{c}{56.5} & 37.1 & \multicolumn{1}{c}{58.9} & 43.4 & 55.3 & 51.3 & \multicolumn{1}{c}{67.7} & 50.1 & \multicolumn{1}{c}{65.9} & 44.6 & 63.0 \\
\multicolumn{1}{c}{L2P\textsuperscript{\textdagger}}  
& 63.4 & \multicolumn{1}{c}{65.1} & 59.0 & \multicolumn{1}{c}{58.2} &  2.6 & 5.6 & 53.9 & \multicolumn{1}{c}{51.6} & 62.9 & \multicolumn{1}{c}{71.4} & 38.7 & 32.2 \\
\multicolumn{1}{c}{FedCLM\textsuperscript{\textdagger}}  
& 58.9 & \multicolumn{1}{c}{69.4} & 57.6 & \multicolumn{1}{c}{67.6} &  44.3 & 57.9 & 66.5 & \multicolumn{1}{c}{77.4} & 61.0 & \multicolumn{1}{c}{71.8} & 48.8 & 63.5 \\
\multicolumn{1}{c}{FedNCM\textsuperscript{\textdagger}}  
& 65.6 & \multicolumn{1}{c}{74.4} & 61.9 & \multicolumn{1}{c}{71.1} &  49.6 & 59.8 & 66.8 & \multicolumn{1}{c}{77.9} & 62.1 & \multicolumn{1}{c}{72.4} & 50.9 & 65.9 \\
\multicolumn{1}{c}{TARGET\textsuperscript{\textdagger}}  
& 60.9 & \multicolumn{1}{c}{71.3} & 58.8 & \multicolumn{1}{c}{69.5} & 45.2 & 56.5 & 66.1 & \multicolumn{1}{c}{77.8} & 60.5 & \multicolumn{1}{c}{71.1} & 51.8 & 65.3 \\
\multicolumn{1}{c}{GLFC\textsuperscript{\textdagger}}  
& 58.2 & \multicolumn{1}{c}{70.4} & 53.7 & \multicolumn{1}{c}{65.9} &  13.1 & 37.7 & - & \multicolumn{1}{c}{-} & - & \multicolumn{1}{c}{-} & - & - \\
\multicolumn{1}{c}{LGA\textsuperscript{\textdagger}}  
& 64.5 & \multicolumn{1}{c}{73.6} & 61.1 & \multicolumn{1}{c}{70.5} &  21.6 & 40.9 & - & \multicolumn{1}{c}{-} & - & \multicolumn{1}{c}{-} & - & - \\
\multicolumn{1}{c}{PILoRA}  
& 69.5 & \multicolumn{1}{c}{78.6} & 65.1 & \multicolumn{1}{c}{74.4} & 54.9 & 62.6 & 68.5 & \multicolumn{1}{c}{78.1} & 63.4 & \multicolumn{1}{c}{73.7} & 54.8 & 67.1 \\
\multicolumn{1}{c}{LoRM} & - & \multicolumn{1}{c}{-} & - & \multicolumn{1}{c}{-} & - & - & 70.5 & \multicolumn{1}{c}{81.1} & 62.4 & \multicolumn{1}{c}{78.2} & 61.0 & 71.8 \\
\multicolumn{1}{c}{InfLoRA\textsuperscript{\textdaggerdbl}} 
& 70.5 & \multicolumn{1}{c}{78.4} & 66.7 & \multicolumn{1}{c}{75.6} & 56.3 & 62.5 & 68.4 & \multicolumn{1}{c}{78.4} & 63.3 & \multicolumn{1}{c}{73.8} &  54.2 & 67.5 \\
\multicolumn{1}{c}{pFedMxF\textsuperscript{\textdaggerdbl}} & 71.3 & \multicolumn{1}{c}{80.7} & 67.4 & \multicolumn{1}{c}{76.2} & 57.0 & 64.9 & 70.2 & \multicolumn{1}{c}{80.3} & 65.6 & \multicolumn{1}{c}{75.2} & 60.5 & 70.5 \\
\multicolumn{1}{c}{\textbf{Fed-TaLoRA}} 
& \textbf{76.6} & \multicolumn{1}{c}{\textbf{84.2}} & \textbf{72.9} & \multicolumn{1}{c}{\textbf{81.0} }& \textbf{66.5} & \textbf{73.2} & \textbf{77.4} & \multicolumn{1}{c}{\textbf{85.5}} & \textbf{76.8} & \multicolumn{1}{c}{\textbf{84.1}} & \textbf{75.3} & \textbf{83.3} \\
\hline
\end{tabular}
}
\end{table*}

\textbf{Non-IID settings}.
In our experiments, we distribute the data of each task across 10 clients, employing the widely adopted non-IID settings: \textit{distribution-based label imbalance} and \textit{quantity-based label imbalance} \splitcite{li2022federated}. 
The former partitions the data based on a Dirichlet distribution controlled by the parameter $\beta$, while the latter ensures that each client contains exactly $\alpha$ classes within each task. 
We assess all methods across 6 different scenarios of non-IID for each dataset. 
Specifically, for distribution-based label imbalance settings, we set $\alpha \in \{2, 4, 6\}$ for CIFAR-100, and $\alpha \in \{4, 8, 12\}$ for Tiny-ImageNet. 
On both datasets, we conduct experiments with $\beta \in \{0.05, 0.1, 0.5\}$.
As shown in Figure \ref{fig:non_iid}, we take CIFA-100 dataset as an example to illustrate the non-IID setting of our experiments.
In the quantity-based label imbalance, we suppose each client only has data samples of $\alpha$ different labels.
We first randomly assign $\alpha$ different label IDs to each client at each task.
Then we randomly and equally divide them into the clients which own the label for the samples of each label.
In the distribution-based label imbalance, we sample $P_c$ from $Dir_N(\beta)$ and allocate a $P_{c, k}$ proportion of the samples of 
class $c$ to the client $k$, where $Dir(\cdot)$ denotes the Dirichlet distribution and $\beta$ is a concentration parameter ($\beta>0$).
The partition is more unbalanced if $\beta$ is set to a smaller value.

\textbf{Baseline.}
We compare the proposed method with existing FCFT methods: TARGET \splitcite{zhang2023targeta}, GLFC \splitcite{dong2022federateda}, LGA \splitcite{dong2024no}, PILoRA \splitcite{guo2024pilora}, LoRM \splitcite{salami2024closed}, and pFedMxF \splitcite{zhang2025pfedmxf}. 
We adopt several representative CIL methods: EWC \splitcite{kirkpatrick2017overcoming}, LwF \splitcite{li2018learninga}, iCaRL \splitcite{rebuffi2017icarl}, L2P \splitcite{wang2022learning}, InfLoRA \splitcite{liang2024inflora}, 
which were originally designed for class-incremental learning, and FL method FedNCM \splitcite{legate2024guiding} and its variant FedCLM \splitcite{legate2024guiding} in the FCFT setting.
For a fair comparison, we tune all methods with the same pre-trained model as ours and fine-tune them using LoRA in the same FCFT settings.

\textbf{Metrics.}
We report two widely adopted metrics in the FCFT setting: Final Average Accuracy ($FAA$) and Average Incremental Accuracy ($AIA$). 
$FAA$ is the average accuracy of all seen classes after learning the final task and $AIA$ is calculated as the average accuracy of all tasks (equivalent to `$A_N$, Avg.' in \splitcite{guo2024pilora,zhang2024slca}). 
To clearly understand the meaning of these metrics, we give their formal definition here.
Let $S_{t, \tau}$ denote the classification accuracy on the $\tau$-th task after training on the $t$-th task.
\begin{equation}
    FAA = A_T, A_t = \frac{1}{t} \sum\limits_{\tau=1}^t S_{t, \tau},
\end{equation}
where $A_t$ denotes the average accuracy up to the $t$-th task and $T$ is the total number of all tasks.
Accordingly, the formal definition of $AIA$ can be formulated as:
\begin{equation}
    AIA = \frac{1}{T} \sum\limits_{t=1}^T A_t.
\end{equation}

\textbf{Implementation.}
We run all experiments of all datasets on two GeForce RTX 3090 GPUs and two GeForce RTX 5090 GPUs using PyTorch \splitcite{paszke2019pytorch}. 
Following previous works \splitcite{zhang2024slca, guo2024pilora} and considering the potential data leakage issue, we evaluate the performance of our method with self-supervised pre-trained weights of DINO \splitcite{caron2021emerging} for ViT-B/16 \splitcite{dosovitskiy2020image}.
An SGD optimizer is used for ours with the same batch size of 64 as baselines. 
Following \splitcite{zhang2024slca}, 
to effectively balance the preservation of pre-trained knowledge stability with the adaptability required for new tasks, 
we use the learning rate of 0.001 for the representation layer and 0.01 for the classification layer.
We initialize 10 local clients, and the number of local training epochs is set to 5.
For each task, 30 rounds of communication are set for all benchmark datasets.
We conduct experiments for 3 times with 3 random seeds (1993, 1996, 1997) and report the averaged results. 
Considering the trade-off between model performance and the number of trainable parameters, inspired by previous efforts in fixed layer select strategy for fine-tuning \splitcite{lee2019would,lee2022surgical},
for CIFAR-100, we train LoRA in the first 4 blocks, middle 4 blocks and last 4 blocks, named \textbf{first}, \textbf{mid} and \textbf{last};
for Tiny-ImageNet, we train LoRA in the first 6 blocks and the last 6 blocks, named \textbf{bottom} and \textbf{top}.
Note that in all experiments, \textbf{first} and \textbf{bottom} are noted as \textbf{Fed-TaLoRA} since the LoRA module embedded in the first block is utilized in \splitcite{guo2024pilora} for a relatively fair comparison. 
Please refer to extensive experiments of the impact of LoRA embedded in the different blocks and layers in Appendix \ref{appendix:lora}. 

\subsection{Experimental Results}

\subsubsection{Results on CIFAR-100 and Tiny-ImageNet}
\label{sec:compare_exps}
\begin{table*}[!ht]
\caption{Results (\%) on Tiny-ImageNet including 10 tasks (20 classes per task) and under different degrees of two non-IID settings.Results marked with $\dagger$ are from \splitcite{guo2024pilora}, and those marked with $\dagger\dagger$ are from \splitcite{zhang2025pfedmxf}.}
\label{tab:tiny}
\centering
\scalebox{1.2}{
\begin{tabular}{cc cc cc cc cc cc c} 
\hline
\multicolumn{1}{c}{\bf{Non-IID}} & \multicolumn{6}{c}{\bf{Quantity-based label imbalance}} & \multicolumn{6}{c}{\bf{Distribution-based label imbalance}} \\  \cline{2-13}
\multicolumn{1}{c}{\bf{Partition}} & \multicolumn{2}{c}{$\alpha=12$} & \multicolumn{2}{c}{$\alpha=8$} & \multicolumn{2}{c}{$\alpha=4$} & \multicolumn{2}{c}{$\beta=0.5$} & \multicolumn{2}{c}{$\beta=0.1$} & \multicolumn{2}{c}{$\beta=0.05$} \\ \cline{2-13}
\bf{Method} & $FAA$ & \multicolumn{1}{c}{$AIA$} & $FAA$ & \multicolumn{1}{c}{$AIA$} & $FAA$ & $AIA$ & $FAA$ & \multicolumn{1}{c}{$AIA$} & $FAA$ & \multicolumn{1}{c}{$AIA$} & \multicolumn{1}{c}{$FAA$} & \multicolumn{1}{c}{$AIA$} \\ 
\hline
\multicolumn{1}{c}{Joint}  
& 83.6 & \multicolumn{1}{c}{-} & 82.9 & \multicolumn{1}{c}{-} & 80.2 & - & 84.3 & \multicolumn{1}{c}{-} & 83.3 & \multicolumn{1}{c}{-} & 82.8 & - \\
\multicolumn{1}{c}{L2P\textsuperscript{\textdagger}}  
& 61.6 & \multicolumn{1}{c}{58.0} & 49.4 & \multicolumn{1}{c}{39.3} &  8.2 & 10.2 & 64.2 & \multicolumn{1}{c}{66.9} & 56.3 & \multicolumn{1}{c}{52.5} &  51.9 & 43.2 \\
\multicolumn{1}{c}{FedCLM\textsuperscript{\textdagger}}  
& 61.6 & \multicolumn{1}{c}{72.4} & 51.8 & \multicolumn{1}{c}{60.3} & 45.8 & 56.9 & 66.5 & \multicolumn{1}{c}{77.4} & 60.4 & \multicolumn{1}{c}{71.0} & 46.7 & 57.8 \\
\multicolumn{1}{c}{FedNCM\textsuperscript{\textdagger}}  
& 71.6 & \multicolumn{1}{c}{81.6} & 69.5 & \multicolumn{1}{c}{79.4} & 57.2 & 64.7 & 73.7 & \multicolumn{1}{c}{81.6} & 70.8 & \multicolumn{1}{c}{80.4} & 68.4 & 78.0 \\
\multicolumn{1}{c}{TARGET\textsuperscript{\textdagger}}  
& 72.6 & \multicolumn{1}{c}{81.6} & 70.3 & \multicolumn{1}{c}{79.6} & 63.8 & 73.5 & 71.6 & \multicolumn{1}{c}{80.9} & 71.0 & \multicolumn{1}{c}{80.1} & 69.3 & 79.1 \\
\multicolumn{1}{c}{GLFC\textsuperscript{\textdagger}}  
& 69.1 & \multicolumn{1}{c}{77.9} & 61.3 & \multicolumn{1}{c}{73.5} & 25.1 & 39.4 & - & \multicolumn{1}{c}{-} & - & \multicolumn{1}{c}{-} &  - & - \\
\multicolumn{1}{c}{LGA\textsuperscript{\textdagger}}  
& 71.3 & \multicolumn{1}{c}{79.4} & 65.8 & \multicolumn{1}{c}{75.3} & 36.7 & 48.8 & - & \multicolumn{1}{c}{-} & - & \multicolumn{1}{c}{-} &  - & - \\
\multicolumn{1}{c}{PILoRA}  
& 74.4 & \multicolumn{1}{c}{81.0} & 74.3 & \multicolumn{1}{c}{80.9} & 70.1 & 77.8 & 74.5 & \multicolumn{1}{c}{80.9} & 74.3 & \multicolumn{1}{c}{81.0} & 73.6 & 80.2 \\
\multicolumn{1}{c}{InfLoRA\textsuperscript{\textdaggerdbl}}  
& - & \multicolumn{1}{c}{-} & - & \multicolumn{1}{c}{-} & - & - & 74.3 & \multicolumn{1}{c}{80.6} & 74.3 & \multicolumn{1}{c}{81.1} & 72.9 & 79.8 \\
\multicolumn{1}{c}{pFedMxF\textsuperscript{\textdaggerdbl}} & - & \multicolumn{1}{c}{-} & - & \multicolumn{1}{c}{-} & - & - & 76.2 & \multicolumn{1}{c}{82.4} & 76.1 & \multicolumn{1}{c}{82.3} & 74.5 & 81.9 \\
\multicolumn{1}{c}{\textbf{Fed-TaLoRA}}  
& \textbf{77.2} & \multicolumn{1}{c}{\textbf{83.0}} & \textbf{76.2} & \multicolumn{1}{c}{\textbf{82.8}} & \textbf{74.1} & \textbf{80.9} & \textbf{78.0} & \multicolumn{1}{c}{\textbf{83.9}} & \textbf{77.0} & \multicolumn{1}{c}{\textbf{\textbf{82.6}}} & \textbf{75.5}  & \textbf{81.9} \\
\hline
\end{tabular}
}
\end{table*}

We evaluate all methods using the $FAA$ and $AIA$ metrics, with comprehensive results presented in Tables \ref{tab:cifar100} and \ref{tab:tiny}.
We demonstrate that Fed-TaLoRA consistently outperforms all baseline methods in various non-IID scenarios.
Among existing FCFT approaches, TARGET maintains relatively stable performance under different degrees of data heterogeneity, exhibiting notable robustness.
In contrast, GLFC and LGA experience significant performance degradation when confronted with extreme quantity-based label imbalance.
Although PILoRA achieves better results than these methods, it still suffers considerable performance decline under severe heterogeneity conditions, highlighting a common limitation among current FCFT methods in handling non-IID data distributions, particularly in scenarios with extreme heterogeneity.
Our method delivers substantial performance improvements over the current SOTA method across both datasets. 
On CIFAR-100, Fed-TaLoRA surpasses PILoRA by margins ranging from 7.1\% $\sim$ 21.5\% in terms of $FAA$.
More impressively, our method maintains the superior performance even under extreme data heterogeneity where competing approaches falter significantly.
On the more complex Tiny-ImageNet dataset, our method consistently outperforms PILoRA by 1.9\% $\sim$ 4.0\% across all evaluated scenarios. 
Additionally, the proposed Fed-TaLoRA delivers superior performance to more recent FCFT methods such as LoRM and pFedMxF.
These results illustrate the efficiency of \textit{task-agnostic adaptation}: directly fine-tuning task-agnostic parameters yields superior efficiency compared to adapting models that have undergone task-specific fine-tuning.

\begin{table}[!ht]
    \centering
        \caption{Results ($\%$) on ImageNet including 10 tasks (100 classes per task).}
        \label{tab:imgn_T10}
        \scalebox{1.1}{
        \begin{tabular}{ccccc} 
        \hline
        \multirow{2}{*}{\bf{Method}} & \multicolumn{2}{c}{$\alpha=60$} & \multicolumn{2}{c}{$\beta=0.5$} \\ \cline{2-5}
         & $FAA$ & $AIA$ & $FAA$ & $AIA$ \\ 
        \hline
        \multicolumn{1}{c}{PILoRA}  
        & 68.0 & 73.6 & 67.3 & 72.8 \\
        \multicolumn{1}{c}{Ours (block 1, w/ MLP)} 
        & 73.3 & 78.9 & 74.5 & 80.6 \\
        \multicolumn{1}{c}{Ours (block 1, w/o MLP)}
        & 72.9 & 78.5 & 74.6 & 80.6 \\
        \multicolumn{1}{c}{Ours (block 1-6, w/ MLP)}  
        & 73.5 & 79.0 & 74.8 & 80.7 \\
        \hline
        \end{tabular}
        }
\end{table}

\begin{table}[!ht]
  \begin{center}
    \centering
        \caption{Ablation Results (\%) on CIFAR-100.}
        \label{tab:albation}
        \scalebox{1.1}{
        \begin{tabular}{ccccc} 
        \hline
        \bf{Partition} & \multicolumn{2}{c}{$\alpha=6$} & \multicolumn{2}{c}{$\beta = 0.5$} \\ \cline{2-5}
        \bf{Method} & $FAA$ & $AIA$ & $FAA$ & $AIA$ \\ 
        \hline
        \multicolumn{1}{c}{w/o \texttt{TaLoRA} and \texttt{ResWU}} & 69.6 & 78.7 & \multicolumn{1}{c}{71.4} & 81.5 \\ 
        \multicolumn{1}{c}{w/o \texttt{TaLoRA} (frozen)}  
        & 69.8 & 78.9 & \multicolumn{1}{c}{71.4} & 81.5 \\
        \multicolumn{1}{c}{w/o \texttt{ResWU}} 
        & 76.5 & 82.9 & \multicolumn{1}{c}{77.4} & 82.8 \\
        \multicolumn{1}{c}{Fed-TaLoRA (fully)}  
        & 78.2 & 86.9 & \multicolumn{1}{c}{79.4} & 87.1 \\
        \multicolumn{1}{c}{\bf{Fed-TaLoRA}}  
        & 76.6 & 84.2 & \multicolumn{1}{c}{77.4} & 85.5 \\
        \hline
        \end{tabular}}
    \end{center}
\end{table}
\subsubsection{Results on ImageNet}

To evaluate our model's scalability to larger datasets, we conducted comprehensive experiments on ImageNet. As shown in Table \ref{tab:imgn_T10}, even our minimal variant—with task-agnostic LoRA embedded only in the first transformer block—significantly outperforms PILoRA. These results demonstrate not only the effectiveness of our approach to complex, large-scale datasets but also highlight promising opportunities for further optimization of the efficiency-performance trade-off. 
The strong performance with such limited parameter adaptation suggests that strategic LoRA deployment can achieve superior results while maintaining minimal resource requirements, reinforcing Fed-TaLoRA's scalability potential for real-world applications with varying computational constraints, since it eliminates the additional costs as the number of tasks grows \splitcite{wu2025sd}.

\subsection{Ablation Study}
\label{sec:ablation}
To evaluate the effect of each component of Fed-TaLoRA, we performed an ablation study on the CIFAR-100 setup of 10 tasks.
Fed-TaLoRA contains two core components, i.e., \texttt{TaLoRA} and \texttt{ResWU}.
We conducted five different experiments as follows:
\begin{enumerate}
    \item Fed-TaLoRA w/o \texttt{TaLoRA} and \texttt{ResWU}: fine-tuning the linear classifiers without applying \texttt{TaLoRA} and \texttt{ResWU} while freezing the entire backbone.
    \item Fed-TaLoRA w/o \texttt{TaLoRA} (frozen): applying \texttt{ResWU} while freezing the entire backbone.
    \item Fed-TaLoRA w/o \texttt{ResWU}: applying \texttt{TaLoRA} without \texttt{ResWU}. 
    \item Fed-TaLoRA (fully): applying \texttt{ResWU} and continually fine-tuning the entire backbone.
    \item Fed-TaLoRA: applying \texttt{TaLoRA} and \texttt{ResWU}.
\end{enumerate}

As shown in Table \ref{tab:albation}, the classification precision $FAA$ and $AIA$ of Fed-TaLoRA w/o \texttt{TaLoRA} and \texttt{ResWU} frozen decrease by 8.4\% and 8.0\%, respectively, compared to the method Fed-TaLoRA (fully fine-tuning) when $\alpha=6$. 
This suggests that the representation ability of the former method is limited, as only the classifier with few parameters is trained while most of the backbone parameters remain unchanged.

In contrast, compared to fully fine-tuning, our method achieves comparable performance by tuning task-agnostic LoRA parameters at significantly lower computational and communication costs.
Furthermore, our ablation studies demonstrate that without the ResWU mechanism, our method's performance significantly degrades, particularly in terms of  $AIA$. 
This underscores the critical role that ResWU plays in enabling accurate model aggregation at the server, thereby enhancing the model's generalization capabilities and robustness when learning across heterogeneous data distributions.

\begin{figure}[!ht]
    \centering
    \subfloat{
    \centering
    \includegraphics[width=0.49\linewidth]{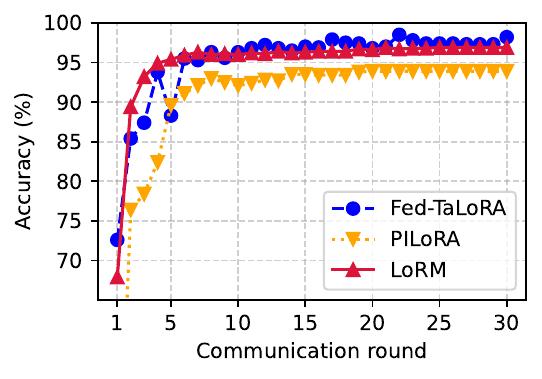}}
    \subfloat{
    \centering
    \includegraphics[width=0.48\linewidth]{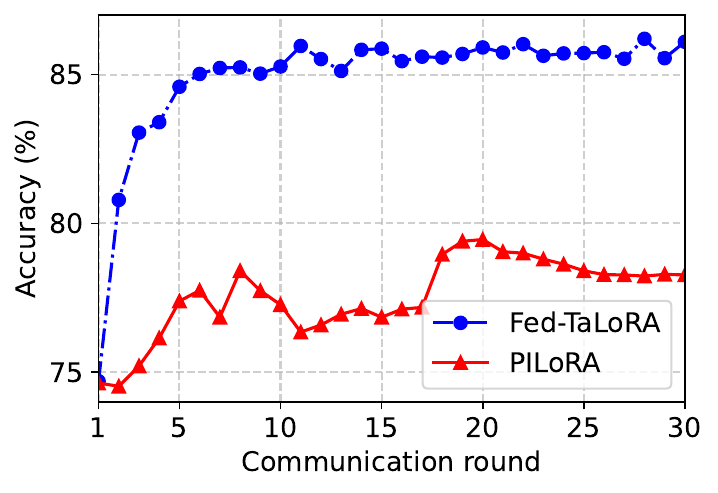}}
    \caption{Speed of convergence on the first task of CIFAR-100 with $\beta=0.5$ (left) and Tiny-ImageNet with $\beta=0.5$ (right).}
    \label{fig:convergence}
\end{figure}

\textit{Rate of convergence.} 
Figure \ref{fig:convergence} illustrates the convergence rates of Fed-TaLoRA and competing FCFT methods on the first task of CIFAR-100 with $\beta=0.5$ (left) and Tiny-ImageNet with $\beta=0.5$ (right).
On CIFAR-100 dataset, the steeper slope of the Fed-TaLoRA curve, relative to those of PILoRA and LoRM, highlights its faster convergence.
This improved convergence is driven by progressive, task-agnostic low-rank fine-tuning combined with accurate model aggregation, further enhanced by the residual weight update mechanism at each communication round.
In contrast, LoRM employs a closed-form solution for merging LoRA modules, while PILoRA relies on simple averaging.
The slower convergence of PILoRA compared to LoRM is likely due to its prototype-based classification strategy, which requires more training rounds to form well-refined prototypes.
On Tiny-ImageNet, Fed-TaLoRA converged at the $6$-th communication round, while PILoRA did not converge until the $25$-th round.

\begin{figure*}[!ht]
    \centering
    \begin{minipage}[b]{0.32\textwidth}
        \includegraphics[width=\textwidth]{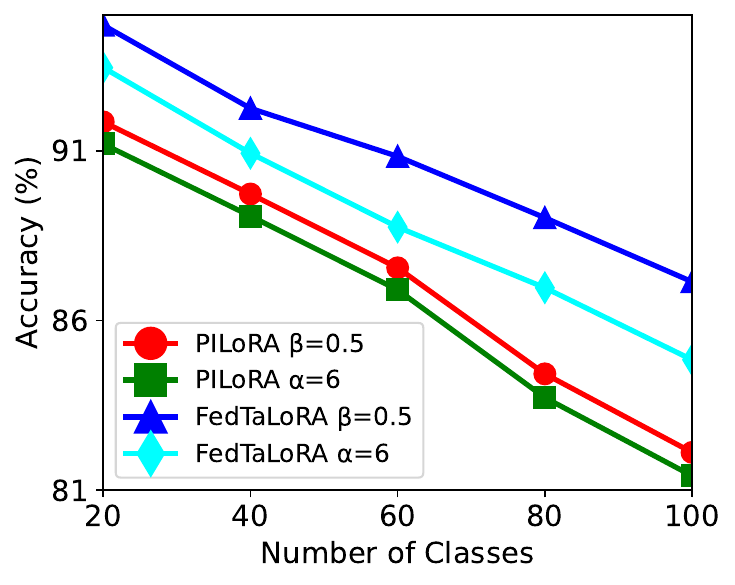}
    \end{minipage}
    \hfill
    \begin{minipage}[b]{0.32\textwidth}
        \includegraphics[width=\textwidth]{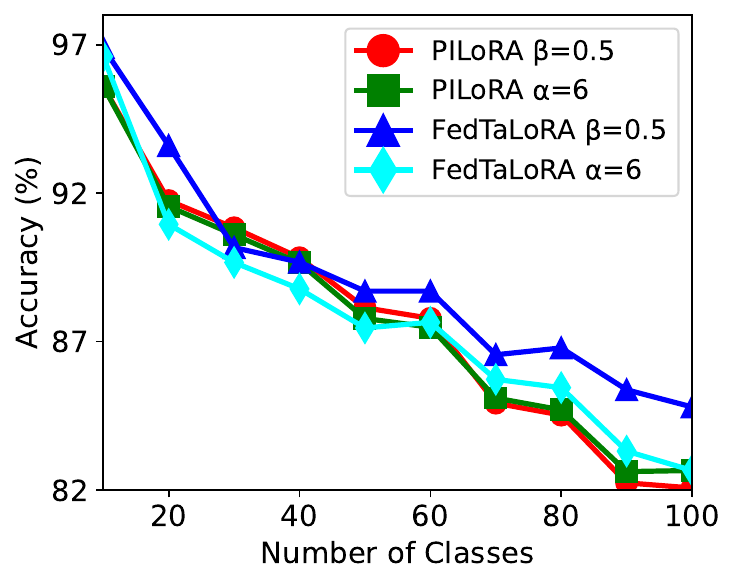}
    \end{minipage}
    \hfill
    \begin{minipage}[b]{0.32\textwidth}
        \includegraphics[width=\textwidth]{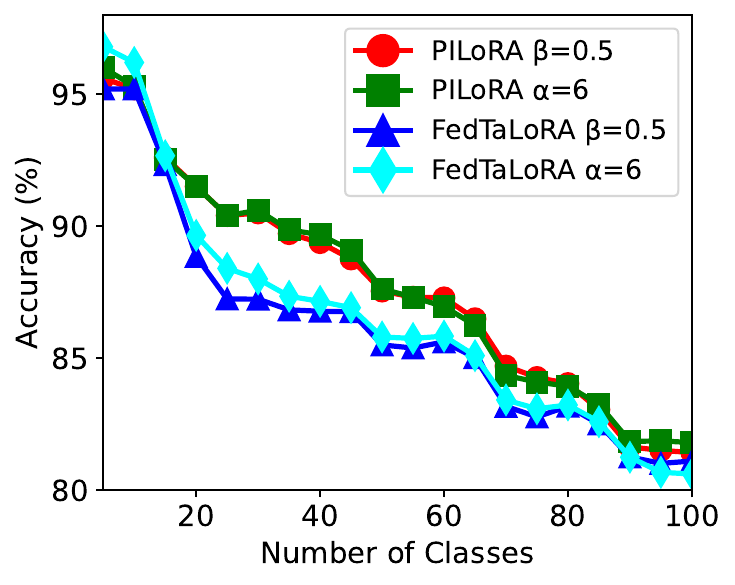}
    \end{minipage}
    \caption{Qualitative analysis of different incremental tasks on ImageNet-Subset when $T=5$ (top), $T=10$ (middle) and $T=20$ (bottom).}
    \label{fig:analysis_mini_various_tasks}
\end{figure*}
\begin{figure*}[!ht]
    \centering
    \begin{minipage}[b]{0.32\textwidth}
        \includegraphics[width=\textwidth]{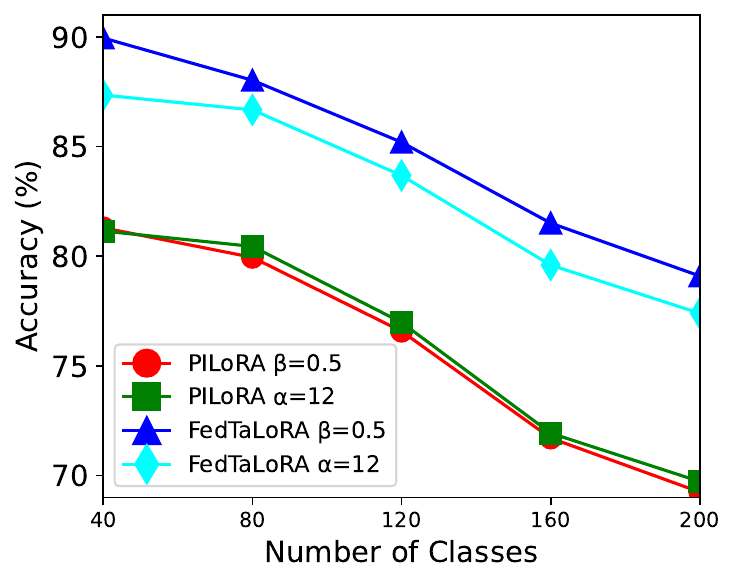}
    \end{minipage}
    \hfill
    \begin{minipage}[b]{0.32\textwidth}
        \includegraphics[width=\textwidth]{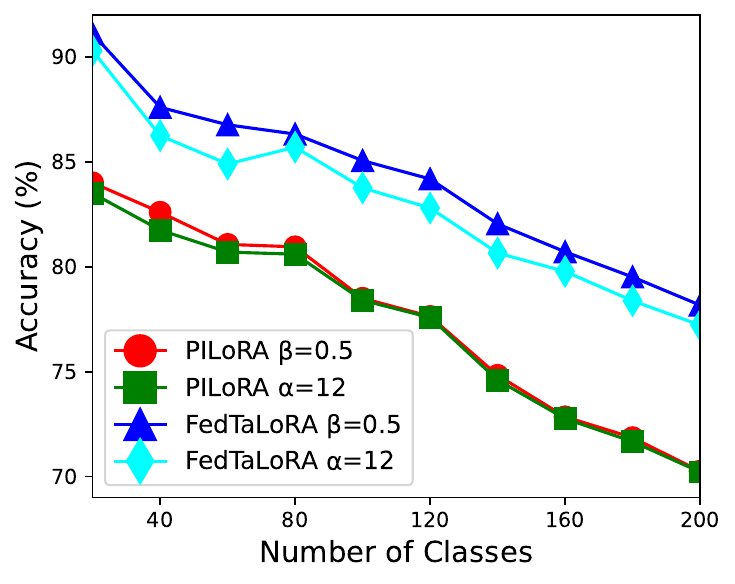}
    \end{minipage}
    \hfill
    \begin{minipage}[b]{0.32\textwidth}
        \includegraphics[width=\textwidth]{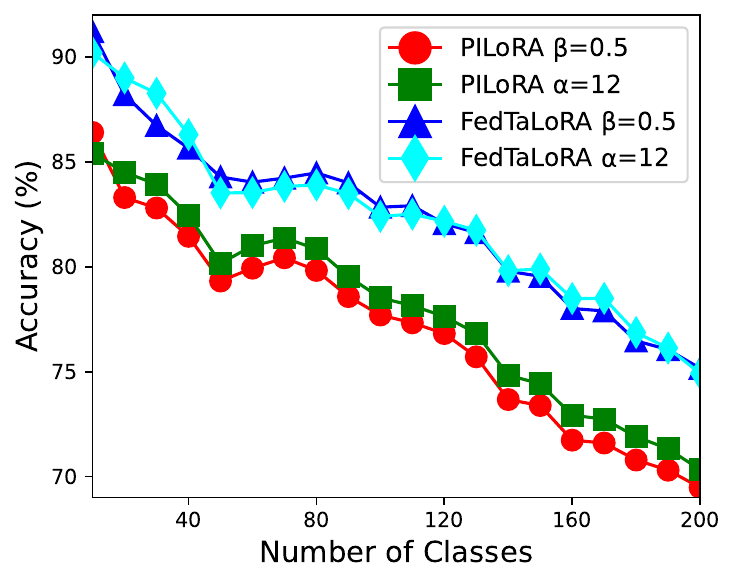}
    \end{minipage}
    \caption{Qualitative analysis of different incremental tasks on Tiny-ImageNet when $T=5$ (top), $T=10$ (middle) and $T=20$ (bottom).}
    \label{fig:analysis_tiny_various_tasks}
\end{figure*}
\subsection{Resource Usage Analysis}
\label{sec:resource_analysis}
In our method, we freeze the pre-trained backbone for all clients and exclusively fine-tune task-shared LoRA parameters throughout the learning process. 
Additionally, we compute the residual weight updates once per communication round to mitigate aggregation inconsistencies in heterogeneous data distributions.

\textbf{(1) Computation and memory efficiency.}
Fed-TaLoRA achieves exceptional resource efficiency across both computation and memory dimensions. 
By fine-tuning only shared task-agnostic LoRA parameters and computing residual weight updates just once per communication round, our method introduces minimal overhead while maintaining superior performance. 
As shown in Table \ref{tab:learnable_params}, Fed-TaLoRA requires fewer trainable parameters than competing methods. 

Unlike PILoRA, which stores separate task-specific parameters and class prototypes for each incremental task, or methods like TARGET, LGA, and GFLC that require growing memory buffers for exemplar storage, Fed-TaLoRA maintains a constant memory footprint regardless of task count. 
This balanced efficiency in both computation and memory makes Fed-TaLoRA particularly well-suited for resource-constrained federated environments where clients have limited capabilities.

\textbf{(2) Communication cost}.
Our empirical evaluation confirms Fed-TaLoRA's optimal balance between communication efficiency and model performance. 
As detailed in Table \ref{tab:communication-cost}, Fed-TaLoRA achieves the lowest communication overhead while simultaneously delivering superior accuracy compared to all baselines. 
Our studies in Appendix \ref{appendix:extended_results} further demonstrate the trade-off of performance and resource costs for \textit{strategic LoRA placement}: strategically embedding task-agnostic LoRA in a reduced subset of Transformer blocks still outperforms current SOTA while reducing additional communication and computational costs. 
These findings underscore the versatility and efficiency of task-agnostic adaptation within Transformer architectures and suggest promising applications beyond FCFT.
\begin{table}[!ht]
\begin{center}
\caption{Comparison of the number of learnable parameters of both datasets.}
\label{tab:learnable_params}
\scalebox{1.2}{
\begin{tabular}{ccc}
    \hline
    \multirow{2}{*}{\bf{Method}} & \multicolumn{2}{c}{\bf{$\#$ Params}} \\ \cline{2-3}
     & CIFAR-100 & Tiny-ImageNet \\ 
    \hline
    PILoRA & 0.21M & 0.29M \\
    \bf{Fed-TaLoRA} & 0.18M & 0.27M \\
    \hline
\end{tabular}
}
\end{center}
\end{table}
\begin{table}[!ht]
\begin{center}
        \centering
        \caption{Comparison of communication cost per round on CIFAR-100, $\alpha=6$.
        The column of $\#$ parameters is sourced from \splitcite{guo2024pilora}.}
        \label{tab:communication-cost}
        \scalebox{1.2}{
        \begin{tabular}{cccc} 
        \hline
        \bf{Method} & \bf{$\#$Params} & $FAA$ & $AIA$  \\ 
        \hline
        \multicolumn{1}{c}{GLFC} 
        & 10.82M & 58.2 & \multicolumn{1}{c}{70.4}  \\
        \multicolumn{1}{c}{LGA} 
        & 10.82M & \multicolumn{1}{c}{64.5} & 73.6 \\
        \multicolumn{1}{c}{TARGET} 
        & 4.64M & 60.9 & \multicolumn{1}{c}{71.3} \\
        \multicolumn{1}{c}{PILoRA}  
        & 0.77M & 69.5 & \multicolumn{1}{c}{78.6} \\
        \multicolumn{1}{c}{\bf{Fed-TaLoRA}}  
        & 0.36M & 76.6 & \multicolumn{1}{c}{84.2} \\
        \hline
        \end{tabular}}
\end{center}
\end{table}
\section{Further Analysis}
\label{sec:fur_analysis}
In this section, we will study the performance on different numbers of incremental tasks, the impact of strategies for accurate aggregation and the number of local clients and the impact of task-agnostic LoRA placements.

\textbf{Qualitative analysis of different numbers of tasks}.
To validate the superior performance of Fed-TaLoRA for $5/10/20$ tasks ($20/10/5$ classes per task), we conducted qualitative analysis on ImageNet-Subset dataset, as shown in Table \ref{tab:subset_extended} and Figure \ref{fig:analysis_mini_various_tasks}.
According to these curves, we can easily observe that Fed-TaLoRA performs much better than the PILoRA method in the two different non-IID settings when $T=5, 10$ but obtain slightly worse than PILoRA when $T=20$. 
We attribute this to coarse Fed-TaLoRA can not capture abundant class representations for each task containing fewer classes, but PILoRA can work better since it extract class-specific prototypes for each class in per task, which is a finer-grained representation process at a much higher resource costs including computation, memory and communication.

Furthermore, we conduct extended experiments on Tiny-ImageNet with $5/10/20$ tasks ($40/20/10$ classes per task).
As shown in Table \ref{tab:tiny_extended} and Figure \ref{fig:analysis_tiny_various_tasks}, Fed-TaLoRA performs much better than PILoRA in all heterogeneous data scenarios.
Therefore, these demonstrate Fed-TaLoRA can largely enhance the model performance combining \texttt{TaLoRA} and \texttt{ResWU} mechanism, further emphasizing the potential of \textit{task-agnostic adaptation} and \textit{post-aggregation model calibration}. 

\begin{table*}[!ht]
    \centering
        \caption{Qualitative analysis on ImageNet-Subset including $5/10/20$ tasks ($20/10/5$ classes per task).}
        \label{tab:subset_extended}
        \scalebox{1.2}{
        \begin{tabular}{ccccccccccccc} 
        \hline
        \multirow{3}{*}{\bf{Method}} & \multicolumn{4}{c}{$T=5$} & \multicolumn{4}{c}{$T=10$} & \multicolumn{4}{c}{$T=20$} \\ \cline{2-13}
        & \multicolumn{2}{c}{$\alpha=6$} & \multicolumn{2}{c}{$\beta=0.5$} & \multicolumn{2}{c}{$\alpha=6$}& \multicolumn{2}{c}{$\beta=0.5$} & \multicolumn{2}{c}{$\alpha=6$} & \multicolumn{2}{c}{$\beta=0.5$} \\ \cline{2-13}
         & $FAA$ & $AIA$ & $FAA$ & $AIA$ & $FAA$ & $AIA$ & $FAA$ & $AIA$ & $FAA$ & $AIA$ & $FAA$ & $AIA$ \\ 
        \hline
        \multicolumn{1}{c}{PILoRA} 
        & 81.4 & 86.5 & 82.1 & 87.1 & 82.7 & 87.8 & 82.1 & 87.8 & 81.8 & 87.7 & 81.9 & 87.8 \\
        \multicolumn{1}{c}{\textbf{Fed-TaLoRA}}  
        & 84.8 & 89.0 & 87.1 & 90.8 & 82.7 & 87.9 & 84.8 & 89.1 & 80.6 & 86.5 & 81.1 & 86.2 \\
        \hline
        \end{tabular}}
\end{table*}

\begin{table*}[!ht]
    \centering
        \caption{Results ($\%$) on Tiny-ImageNet including $5/10/20$ tasks ($40/20/10$ classes per task).}
        \label{tab:tiny_extended}
        \scalebox{1.2}{
        \begin{tabular}{ccccccccccccc} 
        \hline
        \multirow{3}{*}{\bf{Method}} & \multicolumn{4}{c}{$T=5$} & \multicolumn{4}{c}{$T=10$} & \multicolumn{4}{c}{$T=20$} \\ \cline{2-13}
        & \multicolumn{2}{c}{$\alpha=12$} & \multicolumn{2}{c}{$\beta=0.5$} & \multicolumn{2}{c}{$\alpha=12$}& \multicolumn{2}{c}{$\beta=0.5$} & \multicolumn{2}{c}{$\alpha=12$} & \multicolumn{2}{c}{$\beta=0.5$} \\ \cline{2-13}
         & $FAA$ & $AIA$ & $FAA$ & $AIA$ & $FAA$ & $AIA$ & $FAA$ & $AIA$ & $FAA$ & $AIA$ & $FAA$ & $AIA$ \\ 
        \hline
        \multicolumn{1}{c}{PILoRA}  
        & 69.7 & 76.1 & 69.3 & 75.8 & 74.4 & 81.0 & 74.5 & 80.9 & 70.3 & 77.9 & 69.5 & 77.0\\
        \multicolumn{1}{c}{\textbf{Fed-TaLoRA}}  
        & 77.4 & 82.9 & 79.1 & 84.8 & 77.2 & 83.0 & 78.0 & 83.9 & 74.9 & 82.3 & 75.2 & 82.3\\
        \hline
        \end{tabular}}
\end{table*}

\begin{figure}[!ht]
    \centering
    \includegraphics[width=\linewidth]{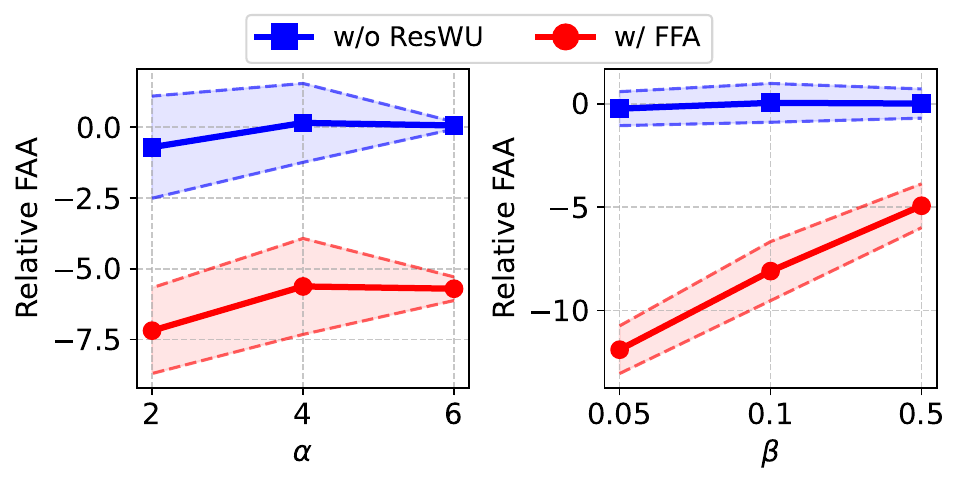}
    \caption{Relative final average accuracy ($\%$) compared to Fed-TaLoRA on CIFAR-100.}
    \label{fig:strategy_accurate_aggregation}
\end{figure}

\textbf{Impact of strategies for accurate aggregation}.
To evaluate approaches for addressing inaccurate aggregation in LoRA-based federated fine-tuning within FCFT context,
we compare our mechanism with alternative strategies. 
We implemented the recent parameter-freezing approach from FFA-LoRA \splitcite{sun2024improvinga} by freezing matrix $\mathbf{A}$ during training (denoted as `w/ FFA' in Figure \ref{fig:strategy_accurate_aggregation}).

Figure \ref{fig:strategy_accurate_aggregation} illustrates the relative final average accuracy of both alternative approaches—Fed-TaLoRA without ResWU (`w/o ResWU') and Fed-TaLoRA with frozen matrix $\mathbf{A}$ (`w/ FFA')—against our complete Fed-TaLoRA implementation.
Across all heterogeneous data scenarios, our model consistently outperforms the model with frozen matrix $\mathbf{A}$. 
This demonstrates that while simply freezing the LoRA matrix offers resource consumption advantages, it significantly degrades model performance in realistic FCFT settings --- contradicting our goal of achieving an optimal balance between performance and resource efficiency in resource-constrained environments.
In contrast, our ResWU mechanism that functions as the \textit{post-aggregation model calibration} consistently improves performance in all evaluation settings, underscoring the importance of model calibration techniques that explicitly address client drift arising from inconsistent local model optimization on heterogeneous datasets \splitcite{zhao2018federated}.
\begin{figure}[!ht]
    \centering
    \includegraphics[width=0.99\linewidth]{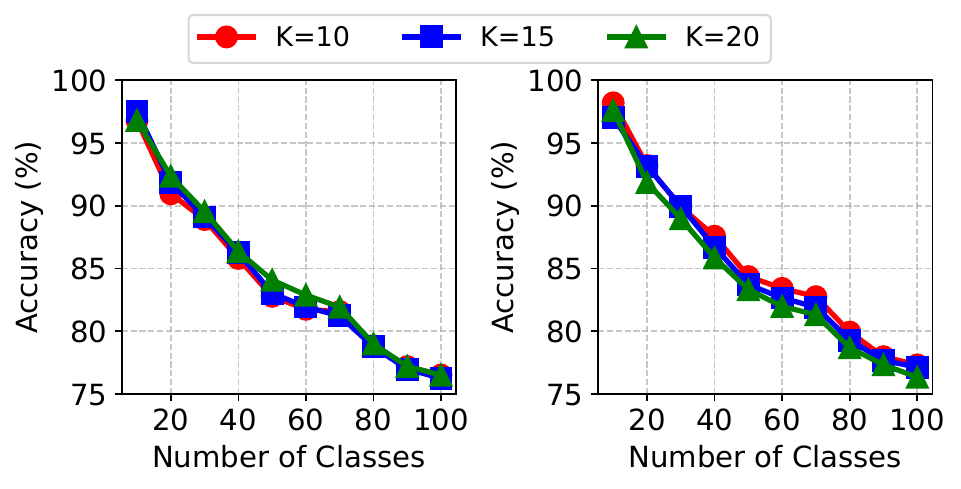}
    \caption{The impact of different number of $K$ on CIFAR-100, $\alpha=6$ (left) and $\beta=0.5$ (right).}
    \label{fig:ablation_K}
\end{figure}

\textbf{Impact of the number of local clients ($K$)}.
\label{appendix:impact_K}
As illustrated in Figure \ref{fig:ablation_K}, we investigate how varying the number of local clients ($K=\{10, 15, 20\}$) affects our model's performance on the CIFAR-100 dataset under different heterogeneity conditions.
For quantity-based label imbalance ($\alpha=6$), we observe a phase-dependent pattern: during the first 5 incremental tasks, configurations with more clients achieve higher accuracy, whereas in the final five tasks, this trend reverses, with fewer clients yielding superior performance. 
This pattern suggests that in early learning stages, the model benefits from the richer feature representations developed from the larger aggregate sample pool available across more clients.
However, as training progresses through additional tasks, the representation enhancement from having more clients reaches diminishing returns, allowing the configuration with fewer clients ($K=10$) to ultimately outperform those with more clients ($K=15$ and $K=20$).
For distribution-based label imbalance ($\beta=0.5$), we observe a slight but consistent decrease in model performance as the number of clients increases, reflecting the greater data heterogeneity introduced by distributing classes across more clients.
Notably, across both heterogeneity scenarios, the variance in performance remains relatively small regardless of client count.
This can be attributed to two key components:
the task-agnostic nature of the trained LoRA parameters and the accurate aggregation enabled by our residual weight update mechanism, which effectively mitigates the challenges of client-count scaling.

\section{Conclusions}
\label{sec:conclusions}
We presented Fed-TaLoRA, a novel and efficient approach for federated continual fine-tuning that addresses the key challenges of catastrophic forgetting, non-IID data heterogeneity, and task-specific parameter inefficiency. Unlike prior methods that rely on task-specific adaptation, Fed-TaLoRA continually fine-tunes a single, shared LoRA module across all clients and tasks. This task-agnostic approach eliminates parameter growth and enables scalable knowledge transfer.
To further ensure accurate aggregation in heterogeneous environments, we propose a residual weight update mechanism that corrects aggregation bias and mitigates client drift. Additionally, we adopt a strategic LoRA placement strategy and dual learning rates for classification and representation layers to improve adaptation efficiency while preserving low communication and computation costs.
We provide a theoretical convergence guarantee under the standard assumptions, then analyze the residual weight update mechanism and prove that it can exactly reconstruct dense aggregation to achieve accurate federated aggregation. 
Extensive experiments on four benchmark datasets demonstrate that Fed-TaLoRA achieves state-of-the-art accuracy with significantly reduced resource usage.

\bibliographystyle{IEEEtran}
\bibliography{references}

\clearpage
\twocolumn[
\begin{center}
\bfseries APPENDIX
\end{center}
\vspace{0.5em}
]
\addcontentsline{toc}{section}{Appendix}

\setcounter{section}{0}
\renewcommand{\thesection}{\Alph{section}}
\renewcommand{\thesubsection}{\thesection.\arabic{subsection}}
\renewcommand{\thesubsubsection}{\thesubsection.\arabic{subsubsection}}

\renewcommand{\theHsection}{app.\Alph{section}}
\renewcommand{\theHsubsection}{\theHsection.\arabic{subsection}}
\section{Proof of the convergence} \label{app:proof_convergence}
In this section, we give the detailed proofs of Lemma~\ref{lem:one_round_full} and Theorem~\ref{thm:stationarity_full} in Section~\ref{sec:theory_analysis}.

\textbf{Lemma 1} (One-Round Progress).
Under Assumptions~\ref{ass:smooth_full}--\ref{ass:hetero_full}, choose $\eta\le\frac{1}{4L\varepsilon}$. For any round $r$,
\begin{align*}
    \mathbb{E}\big[\mathcal{L}(\theta^{r+1})\mid\theta^{r}\big]
    &\le \mathcal{L}(\theta^{r})
      -\frac{\eta\varepsilon}{2}\big\|\nabla\mathcal{L}(\theta^{r})\big\|^2 
      + \eta^2 L \varepsilon\big(\sigma^2 + \varepsilon\delta^2\big),
\end{align*}
where the expectation is taken over mini-batch randomness.

\begin{proof}
Condition on $\theta^{r}$. Under full client participation, each client $k$ performs $\varepsilon$ SGD steps:
$\theta_{k}^{r,e+1}=\theta_{k}^{r,e}-\eta g_k^{r,e}$,
$e=0,\dots,\varepsilon-1$, where $g_k^{r,e}$ is the stochastic gradient at step $e$.
Because ResWU reconstructs the weighted dense aggregation target (Appendix~\ref{subsec:reswu_analysis}), the aggregated update equals
\[
\Delta^{r}=\theta^{r+1}-\theta^{r}
=
-\eta \sum_{k=1}^{K}\omega_k \sum_{e=0}^{\varepsilon-1} g_k^{r,e}.
\]
By $L$-smoothness,
\begin{equation}
\label{eq:smoothness_step}
\mathcal{L}(\theta^{r+1}) \le \mathcal{L}(\theta^{r})
+ \left\langle \nabla\mathcal{L}(\theta^{r}), \Delta^{r}\right\rangle
+ \frac{L}{2} \|\Delta^{r}\|^2.
\end{equation}

\paragraph{Bounding the expected descent term.}
Taking conditional expectation and using Assumption~\ref{ass:variance_full},
\begin{align}
\label{eq:expected_update}
\mathbb{E}[\Delta^{r}\mid\theta^{r}]
&= -\eta \sum_{k=1}^{K}\omega_k 
        \sum_{e=0}^{\varepsilon-1} \mathbb{E}[g_k^{r,e}\mid\theta^{r}] \nonumber \\
&= -\eta \sum_{k=1}^{K}\omega_k 
        \sum_{e=0}^{\varepsilon-1} \mathbb{E}\big[\nabla\mathcal{L}_k(\theta_{k}^{r,e}) \mid \theta^{r}\big].
\end{align}
Add and subtract $\nabla\mathcal{L}_k(\theta^{r})$ in each term:
\begin{align*}
\left\langle \nabla\mathcal{L}(\theta^{r}), \mathbb{E}[\Delta^{r}\mid\theta^{r}] \right\rangle
&= -\eta\varepsilon \big\|\nabla\mathcal{L}(\theta^{r})\big\|^2 + R_1,
\end{align*}
with
\[
R_1 = -\eta \sum_{k=1}^{K}\omega_k \sum_{e=0}^{\varepsilon-1}
      \left\langle \nabla\mathcal{L}(\theta^{r}),
      \mathbb{E}\big[\nabla\mathcal{L}_k(\theta_{k}^{r,e}) - \nabla\mathcal{L}_k(\theta^{r}) \mid \theta^{r}\big]
      \right\rangle.
\]
By Cauchy--Schwarz 
and Assumption~\ref{ass:smooth_full},
\[
\|\nabla\mathcal{L}_k(\theta_{k}^{r,e}) - \nabla\mathcal{L}_k(\theta^{r})\|
\le L \|\theta_{k}^{r,e} - \theta^{r}\|.
\]
Unrolling the recursion gives
\[
\theta_{k}^{r,e}-\theta^{r} = -\eta \sum_{t=0}^{e-1} g_k^{r,t}.
\]
At this point, directly bounding
$\sum_{e=0}^{\varepsilon-1}\mathbb{E}[\|\theta_{k}^{r,e}-\theta^{r}\|\mid\theta^{r}]$
from
$\mathbb{E}[\|\theta_{k}^{r,e}-\theta^{r}\|^2\mid\theta^{r}]$
would introduce an unnecessary looseness in the dependence on $\varepsilon$.
Therefore, we invoke the standard local-drift estimate used in nonconvex analyses of full-participation local SGD/FedAvg under Assumptions~\ref{ass:smooth_full}--\ref{ass:hetero_full} \splitcite{li2020federated,khaled2020tighter,fallah2020personalized}, which yields
\begin{align}
\eta L \|\nabla\mathcal{L}(\theta^{r})\|
\sum_{k=1}^{K}\omega_k
\sum_{e=0}^{\varepsilon-1}
\mathbb{E}\big[\|\theta_{k}^{r,e}-\theta^{r}\|\mid\theta^{r}\big]
\le
\eta^2 L \varepsilon(\sigma^2+\varepsilon\delta^2) \nonumber \\
+
\frac{\eta\varepsilon}{4}\|\nabla\mathcal{L}(\theta^{r})\|^2.
\end{align}
Hence,
\[
|R_1|
\le
\eta^2 L \varepsilon(\sigma^2+\varepsilon\delta^2)
+
\frac{\eta\varepsilon}{4}\|\nabla\mathcal{L}(\theta^{r})\|^2.
\]

\paragraph{Bounding the quadratic term.}
We now control $\mathbb{E}\|\Delta^{r}\|^2$. Using the standard second-moment estimate for full-participation local SGD under Assumptions~\ref{ass:variance_full} and \ref{ass:hetero_full} \splitcite{li2020federated,khaled2020tighter,fallah2020personalized}, we have
\[
\mathbb{E}\big[\|\Delta^{r}\|^2 \mid \theta^{r}\big]
\le
2\eta^2 \varepsilon
\left(
\sigma^2+\varepsilon\big(\|\nabla\mathcal{L}(\theta^{r})\|^2+\delta^2\big)
\right).
\]
Therefore,
\[
\frac{L}{2}\mathbb{E}\big[\|\Delta^{r}\|^2 \mid \theta^{r}\big]
\le
\eta^2L\varepsilon(\sigma^2+\varepsilon\delta^2)
+
\frac{\eta\varepsilon}{4}\|\nabla\mathcal{L}(\theta^{r})\|^2,
\]
where $\eta\le\frac{1}{4L\varepsilon}$ is used to absorb the gradient term.

\paragraph{Putting everything together.}
Combining the bound on
$\langle\nabla\mathcal{L}(\theta^{r}), \mathbb{E}[\Delta^{r}\mid\theta^{r}]\rangle$
with the estimate of
$\mathbb{E}[\|\Delta^{r}\|^2\mid\theta^{r}]$,
and substituting both into \eqref{eq:smoothness_step}, we obtain
\[
\mathbb{E}\big[\mathcal{L}(\theta^{r+1})\mid\theta^{r}\big]
\le
\mathcal{L}(\theta^{r})
-\frac{\eta\varepsilon}{2}\|\nabla\mathcal{L}(\theta^{r})\|^2
+\eta^2L\varepsilon(\sigma^2+\varepsilon\delta^2),
\]
where the condition $\eta\le \frac{1}{4L\varepsilon}$ is used to absorb the gradient terms. This proves the claim.
\end{proof}

\textbf{Theorem 1} (Stationarity Rate).
Under Assumptions~\ref{ass:smooth_full}--\ref{ass:hetero_full} and $\eta\le\frac{1}{4L\varepsilon}$, running Fed-TaLoRA for $R$ rounds yields
\[
\frac{1}{R}\sum_{r=0}^{R-1}
\mathbb{E}\big[\|\nabla\mathcal{L}(\theta^{r})\|^2\big]
\le 
\frac{2(\mathcal{L}(\theta^{0})-\mathcal{L}^\star)}{\eta\varepsilon R}
+ 2\eta L(\sigma^2+\varepsilon\delta^2),
\]
where $\mathcal{L}^\star$ is the infimum of $\mathcal{L}$.

\begin{proof}
Taking total expectation in Lemma~\ref{lem:one_round_full} and summing over $r=0,\dots,R-1$ gives
\[
\mathbb{E}[\mathcal{L}(\theta^{R})]-\mathcal{L}(\theta^{0})
\le -\frac{\eta\varepsilon}{2}
\sum_{r=0}^{R-1}
\mathbb{E}\|\nabla\mathcal{L}(\theta^{r})\|^2
+ \eta^2 L \varepsilon R (\sigma^2 + \varepsilon \delta^2).
\]
Since $\mathcal{L}(\theta^{R})\ge \mathcal{L}^\star$, rearranging the above inequality and dividing both sides by $\eta\varepsilon R/2$ yields the stated bound. The telescoping argument follows the standard template in \splitcite{li2020federated,khaled2020tighter}.
\end{proof}

\section{LoRA Aggregation Analysis via Residual Weight Update}
\label{subsec:reswu_analysis}
To analyze the aggregation behavior of low-rank adapters, recall that a fully fine-tuned model is aggregated as
\begin{equation}
\label{eq:fft_repeat}
\mathbf{W}=\sum_{k=1}^{K}\omega_k\mathbf{W}_k,
\qquad
\sum_{k=1}^{K}\omega_k=1.
\end{equation}
When only the LoRA factors are trained, the corresponding weighted dense target is
\begin{equation}
\label{eq:lora_dense_repeat}
\mathbf{W}_g^*
=
\mathbf{W}_0+\sum_{k=1}^{K}\omega_k\mathbf{B}_k\mathbf{A}_k.
\end{equation}
In practice, to keep the communication footprint low, each client transmits only its low-rank factors
$\mathbf{B}_k\in\mathbb{R}^{d\times r}$ and $\mathbf{A}_k\in\mathbb{R}^{r\times d'}$,
and the server averages them independently:
\begin{equation}
\label{eq:lora_naive_repeat}
\hat{\mathbf{W}}_g
=
\mathbf{W}_0+\mathbf{B}\mathbf{A},
\qquad
\mathbf{B}=\sum_{k=1}^{K}\omega_k\mathbf{B}_k,
\quad
\mathbf{A}=\sum_{k=1}^{K}\omega_k\mathbf{A}_k.
\end{equation}
The discrepancy is therefore
\begin{align}
\label{eq:lora_bias}
\mathbf{W}_{res}
&=
\sum_{k=1}^{K}\omega_k\mathbf{B}_k\mathbf{A}_k-\mathbf{B}\mathbf{A}
\nonumber\\
&=
\sum_{k=1}^{K}\omega_k\mathbf{B}_k\mathbf{A}_k
-
\sum_{k=1}^{K}\sum_{\ell=1}^{K}\omega_k\omega_{\ell}\mathbf{B}_k\mathbf{A}_{\ell}
\nonumber\\
&= \frac{1}{2}\sum_{k=1}^{K}\sum_{\ell=1}^{K}\omega_k\omega_\ell
   \Big[\mathbf{B}_k\mathbf{A}_k - \mathbf{B}_k\mathbf{A}_\ell
        + \mathbf{B}_\ell\mathbf{A}_\ell - \mathbf{B}_\ell\mathbf{A}_k\Big] \nonumber\\
&=
\frac{1}{2}\sum_{k=1}^{K}\sum_{\ell=1}^{K}\omega_k\omega_{\ell}
(\mathbf{B}_k-\mathbf{B}_{\ell})(\mathbf{A}_k-\mathbf{A}_{\ell}).
\end{align}
where the second line expands ${\mathbf{B}}{\mathbf{A}}$, the third line averages the expression with its $k\leftrightarrow\ell$ counterpart, and the fourth line groups the terms into difference products. 
Equation~\eqref{eq:lora_bias} shows that factor-wise LoRA averaging introduces an additional bias term whenever different clients learn different low-rank updates: 
whenever clients learn distinct low-rank subspaces, $\mathbf{W}_{res}\neq\mathbf{0}$ and the naive update \eqref{eq:lora_naive_repeat} diverges from the dense target \eqref{eq:lora_dense_repeat}, causing client drift \splitcite{babakniya2023sloraa,sun2024improvinga}.

\textbf{Comparison with existing fixes.}
Three prevalent remedies have been explored in prior work.  
(1) \emph{Dense aggregation} explicitly transmits $\mathbf{B}_k\mathbf{A}_k$ to the server and averages them as in \eqref{eq:lora_dense_repeat}. 
This eliminates bias but destroys the communication advantage of LoRA because each client must upload a $d\times d'$ matrix.  
(2) \emph{Factor freezing} keeps one factor (typically $\mathbf{A}$) fixed across all clients \splitcite{sun2024improvinga,zhu2024asymmetry,hao2024flora}. 
The discrepancy then vanishes, but expressivity is reduced and accuracy degrades in practice.  
(3) \emph{Stacking-based aggregation} concatenates multiple adapters before averaging \splitcite{wang2024flora}, preserving flexibility at the price of inflated message size. 
Fed-TaLoRA aims to retain both efficiency and flexibility without introducing bias.

\textbf{Residual Weight Update.}
We use the residual in Eq.~\eqref{eq:lora_bias} to correct the server-side dense model:
\begin{equation}
\label{eq:reswu_update_repeat}
\mathbf{W}_g
=
\mathbf{W}_0+\mathbf{W}_{res}+\mathbf{B}\mathbf{A}.
\end{equation}
Substituting Eq.~\eqref{eq:lora_bias} into Eq.~\eqref{eq:reswu_update_repeat} gives
\[
\mathbf{W}_g
=
\mathbf{W}_0+\sum_{k=1}^{K}\omega_k\mathbf{B}_k\mathbf{A}_k,
\]
which exactly matches the weighted dense target in Eq.~\eqref{eq:lora_dense_repeat}.
Therefore, ResWU removes the factor-wise LoRA aggregation bias while preserving low-rank communication.

Note that the proposed ResWU is applied here to ViT-based FCFT, which is orthogonal to the recent study \splitcite{singhal2025fedex} on language models in conventional federated learning.

\section{Extended Experiments}
In this section, we study the impact of task-agnostic LoRA placements.

\label{appendix:extended_results}
\label{appendix:lora}
It is well-known that embedding LoRA in different blocks has varying effects on model representation, thereby influencing the final model performance.
Next, we will investigate the impact of task-agnostic LoRA placements across different transformer blocks and whether to incorporate LoRA in FFN (also called MLP) layers.

\begin{figure*}[!ht]
    \centering
    \subfloat{
    \centering
    \includegraphics[width=0.47\linewidth]{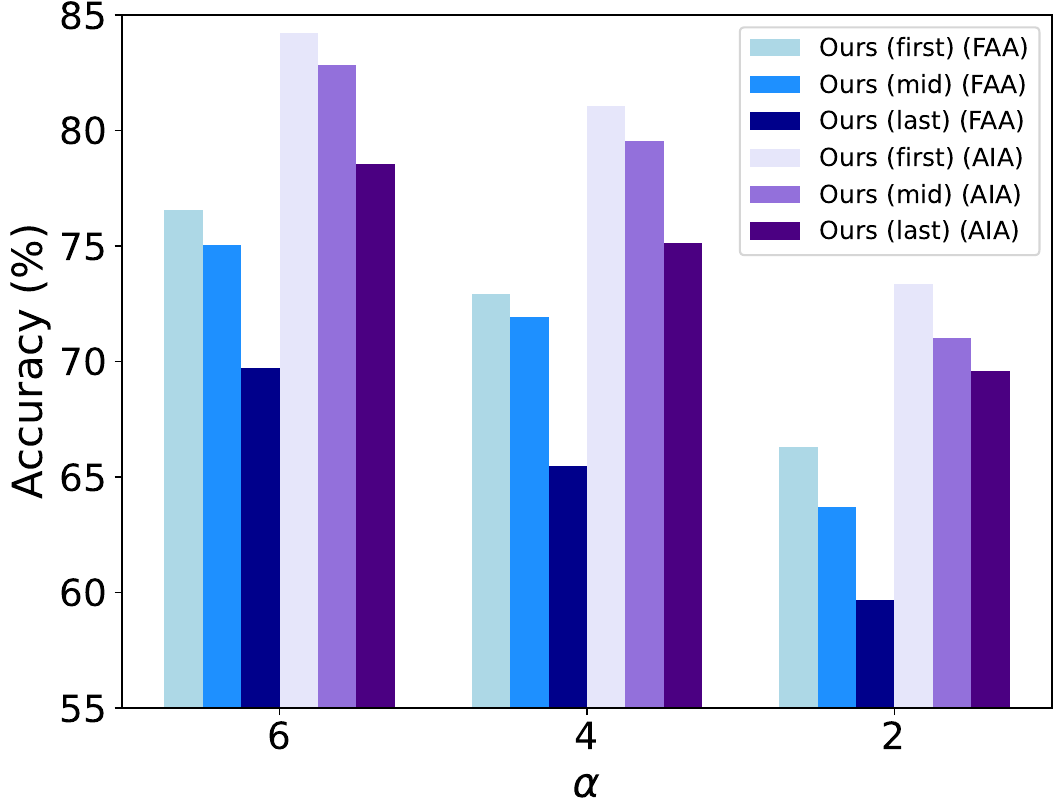}}
    \hfill
    \subfloat{
    \centering
    \includegraphics[width=0.48\linewidth]{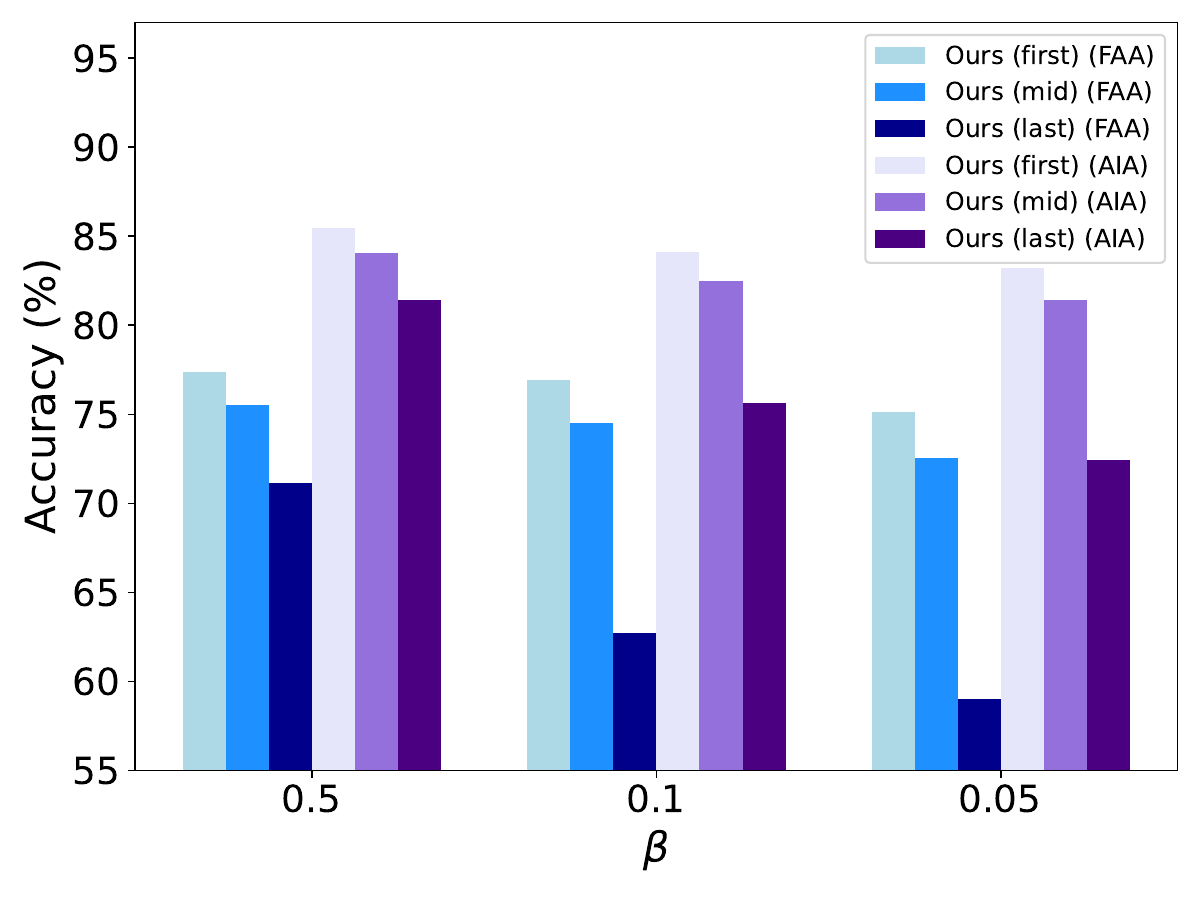}}
    \caption{Performance of LoRA embedded in different blocks for CIFAR-100 dataset.}
    \label{fig:lora_in_differnt_blocks_cifar100}
\end{figure*}

\begin{figure*}[!ht]
    \centering
    \subfloat{
    \centering
    \includegraphics[width=0.47\linewidth]{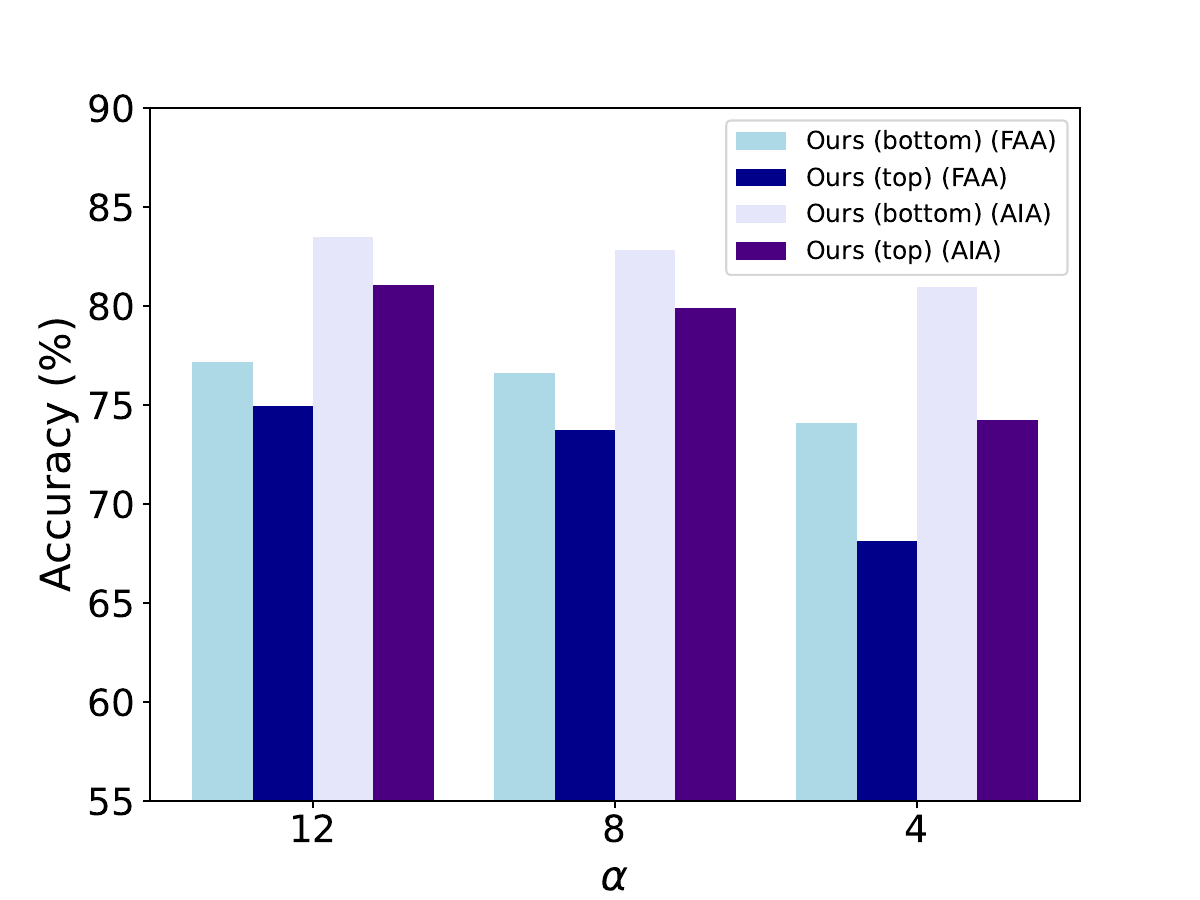}
    }
    \hfill
    \subfloat{
    \centering
    \includegraphics[width=0.48\linewidth]{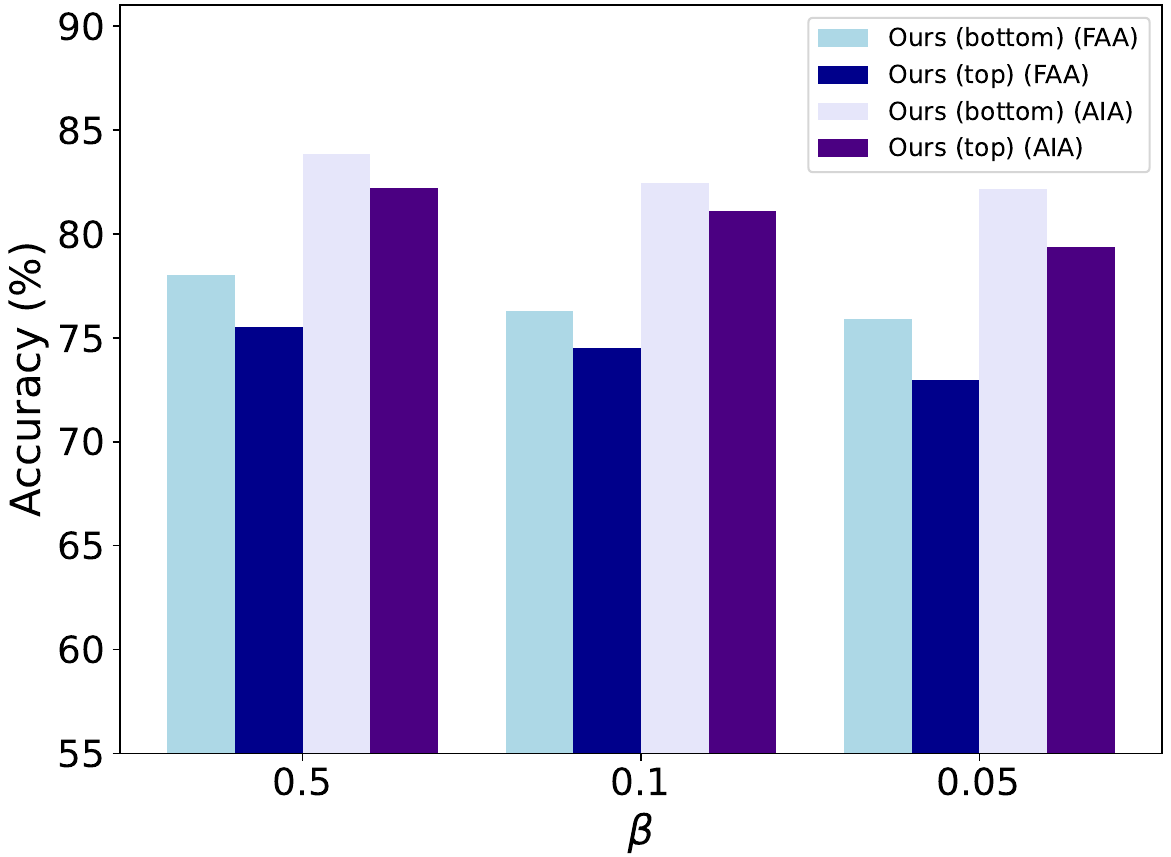}
    }
    \caption{Performance of LoRA embedded in different blocks for Tiny-ImageNet dataset.}
    \label{fig:lora_in_differnt_blocks_tiny}
\end{figure*}

\begin{table}[!ht]
\begin{center}
    \centering
        \caption{Results of different task-agnostic LoRA placements on CIFAR-100.}
        \label{tab:lora_in_which_block_cifar}
        \scalebox{1.1}{
        \begin{tabular}{cc c c c} 
        \hline
        \multirow{2}{*}{\bf{Method}} & \multicolumn{2}{c}{$\alpha=6$} & \multicolumn{2}{c}{$\beta = 0.5$} \\ \cline{2-5}
        & $FAA$ & $AIA$ & $FAA$ & $AIA$ \\ 
        \hline
        \multicolumn{1}{c}{Ours (block 1, w/ MLP)}  
        & 69.1 & 78.3 & \multicolumn{1}{c}{71.0} & 81.2 \\
        \multicolumn{1}{c}{Ours (block 1, w/o MLP)} 
        & 70.7 & 79.5 & \multicolumn{1}{c}{72.1} & 82.0 \\
        \multicolumn{1}{c}{Ours (block 1-4, w/ MLP)}  
        & 76.6 & 84.2 & \multicolumn{1}{c}{77.4} & 85.5 \\
        \multicolumn{1}{c}{Ours (block 1-4, w/o MLP)}
        & 76.4 & 83.9 & 77.3 & 85.4 \\
        \multicolumn{1}{c}{Ours (block 1-12, w/ MLP)}  
        & 76.7 & 84.2 & \multicolumn{1}{c}{77.6} & 85.7 \\
        \hline
        \end{tabular}}
\end{center}
\end{table}
\begin{table}[!ht]
    \centering
        \caption{Results of different task-agnostic LoRA placements on Tiny-ImageNet.}
        \label{tab:lora_in_which_block_tiny}
        \scalebox{1.1}{
        \begin{tabular}{cc c c c} 
        \hline
        \multirow{2}{*}{\bf{Method}} & \multicolumn{2}{c}{$\alpha=12$} & \multicolumn{2}{c}{$\beta = 0.5$} \\ \cline{2-5}
         & $FAA$ & $AIA$ & $FAA$ & $AIA$ \\ 
        \hline
        \multicolumn{1}{c}{Ours (block 1, w/ MLP)}
        & 72.9 & 80.6 & \multicolumn{1}{c}{74.4}  & 81.7 \\
        \multicolumn{1}{c}{Ours (block 1, w/o MLP)}
        & 73.3 & 80.3 & \multicolumn{1}{c}{74.5}  & 81.8 \\
        \multicolumn{1}{c}{Ours (block 1-4, w/ MLP)}
        & 77.1 & 83.0 & \multicolumn{1}{c}{77.8} & 84.0 \\
        \multicolumn{1}{c}{Ours (block 1-4, w/o MLP)} 
        & 77.2 & 82.8 & \multicolumn{1}{c}{77.3} & 83.6 \\
        \multicolumn{1}{c}{Ours (block 1-6, w/ MLP)}  
        & 77.2 & 83.0 & \multicolumn{1}{c}{78.0} & 83.9 \\
        \multicolumn{1}{c}{Ours (block 1-12, w/ MLP)}  
        & 78.0 & 84.0 & \multicolumn{1}{c}{78.2} & 84.2 \\
        \hline
        \end{tabular}}
        
\end{table}
\textit{(1) Exploration of task-agnostic LoRA embedded in specific blocks}.
To evaluate how task-agnostic LoRA placement affects model performance, we conducted experiments with different configurations. 
For CIFAR-100, we compared three placement strategies: the first 4 blocks (denoted as \textbf{first}), middle 4 blocks (\textbf{mid}), and last 4 blocks (\textbf{last}). 
Results are presented in Figure \ref{fig:lora_in_differnt_blocks_cifar100}.
Similarly, for Tiny-ImageNet, we compared two strategies: the first 6 blocks (\textbf{bottom}) and the last 6 blocks (\textbf{top}), with results shown in Figure \ref{fig:lora_in_differnt_blocks_tiny}.
Our findings consistently demonstrate that models with task-agnostic LoRA placed in the lower transformer blocks (\textbf{first} for CIFAR-100 and \textbf{bottom} for Tiny-ImageNet) achieve superior performance compared to other configurations. 
Considering the trade-off between performance and resource usage analysis presented in Section \ref{sec:resource_analysis}, we conclude that these lower block placement strategies represent the optimal design choice for their respective datasets, balancing model quality, computational cost, and memory consumption during inference.

\textit{(2) Exploration of LoRA placement in MLP layers}.
\label{appendix:lora_mlp}
To investigate the optimal placement of task-agnostic LoRA parameters in attention and MLP layers, we conducted comprehensive experiments across CIFAR-100 and Tiny-ImageNet datasets. 
Tables \ref{tab:lora_in_which_block_cifar} and \ref{tab:lora_in_which_block_tiny} present our results, demonstrating that our approach of fine-tuning LoRA parameters in the first transformer block (denoted as \textit{Ours (block 1, w/ MLP)}) marginally outperforms the current PILoRA method, with detailed performance metrics available in Table \ref{tab:cifar100}.

Our experiments reveal intriguing patterns regarding MLP layer inclusion.
When restricting LoRA to only the first block, excluding MLP layers yields superior performance on both datasets. 
However, when extending LoRA across block 1-4, including MLP layers consistently delivers better results.
We attribute these observations to two key factors: 
\textbf{(1)} When fine-tuning a limited parameter set (as in \textit{Ours (block 1, w/ MLP)} and \textit{Ours (block 1, w/o MLP)}, attention layer weights contribute more significantly to model performance than MLP layer weights;
\textbf{(2)} With an expanded parameter space (comparing \textit{Ours (block 1-4, w/ MLP)} and \textit{Ours (block 1-4, w/o MLP)} to their block 1 counterparts), the inclusion of MLP layers provides complementary benefits that enhance multi-layer LoRA fine-tuning effectiveness.

Further evidence from Figures \ref{fig:lora_in_differnt_blocks_cifar100} and \ref{fig:lora_in_differnt_blocks_tiny} suggests a hierarchical feature extraction mechanism, where lower blocks capture coarser features while upper blocks extract finer details.
Notably, our experiments confirm that fine-tuning LoRA across all transformer blocks consistently achieves optimal performance, reinforcing previous findings by \splitcite{fomenko2024note,lin2024tracking} that involving more critical parameters in Vision Transformers during training yields superior results, albeit at increased computational and communication costs.

\begin{figure}[!ht]
    \centering
    \begin{minipage}[b]{0.24\textwidth}
        \includegraphics[width=\textwidth]{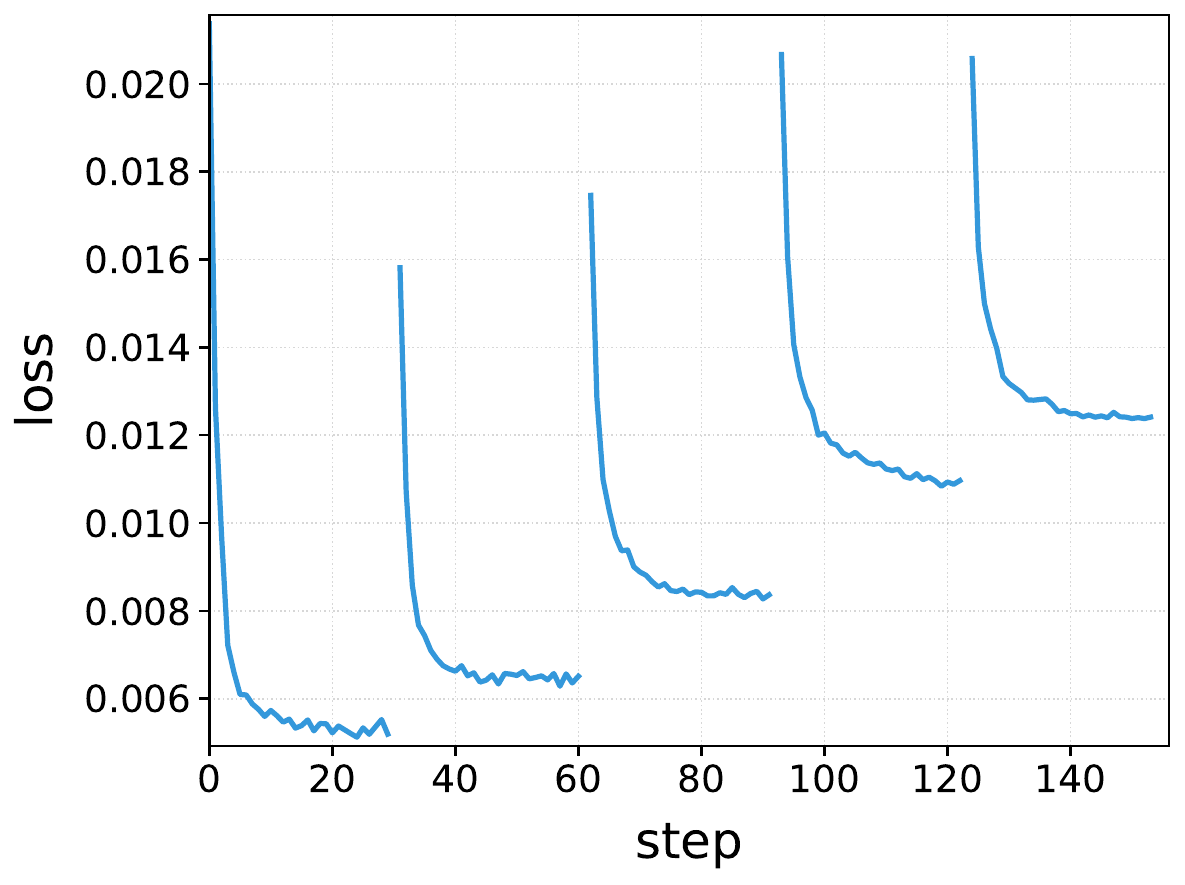}
    \end{minipage}
    \hfill
    \begin{minipage}[b]{0.24\textwidth}
        \includegraphics[width=\textwidth]{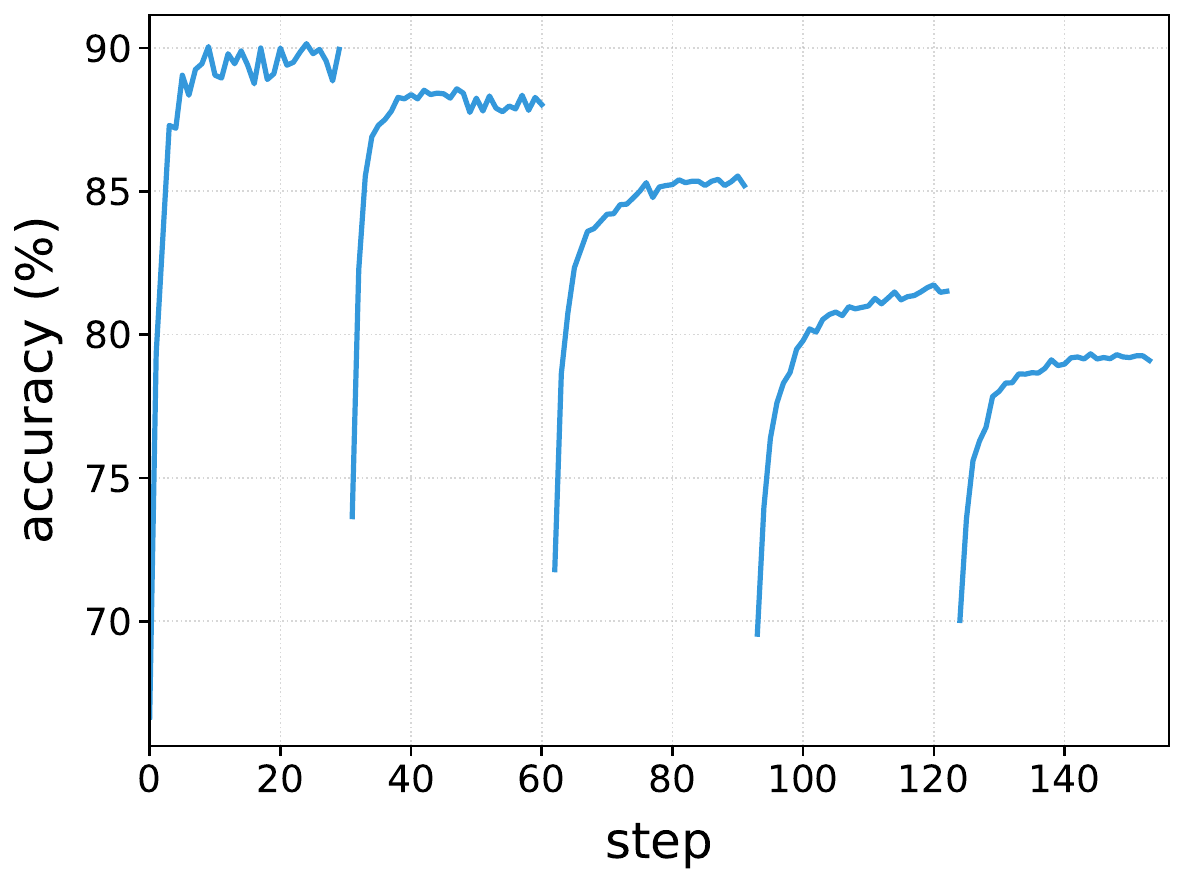}
    \end{minipage}
    \caption{Training curves for $T=5$ on Tiny-ImageNet when $\alpha=12$.}
    \label{fig:curve_T5}
\end{figure}

\begin{figure}[!ht]
    \centering
    \begin{minipage}[b]{0.24\textwidth}
        \includegraphics[width=\textwidth]{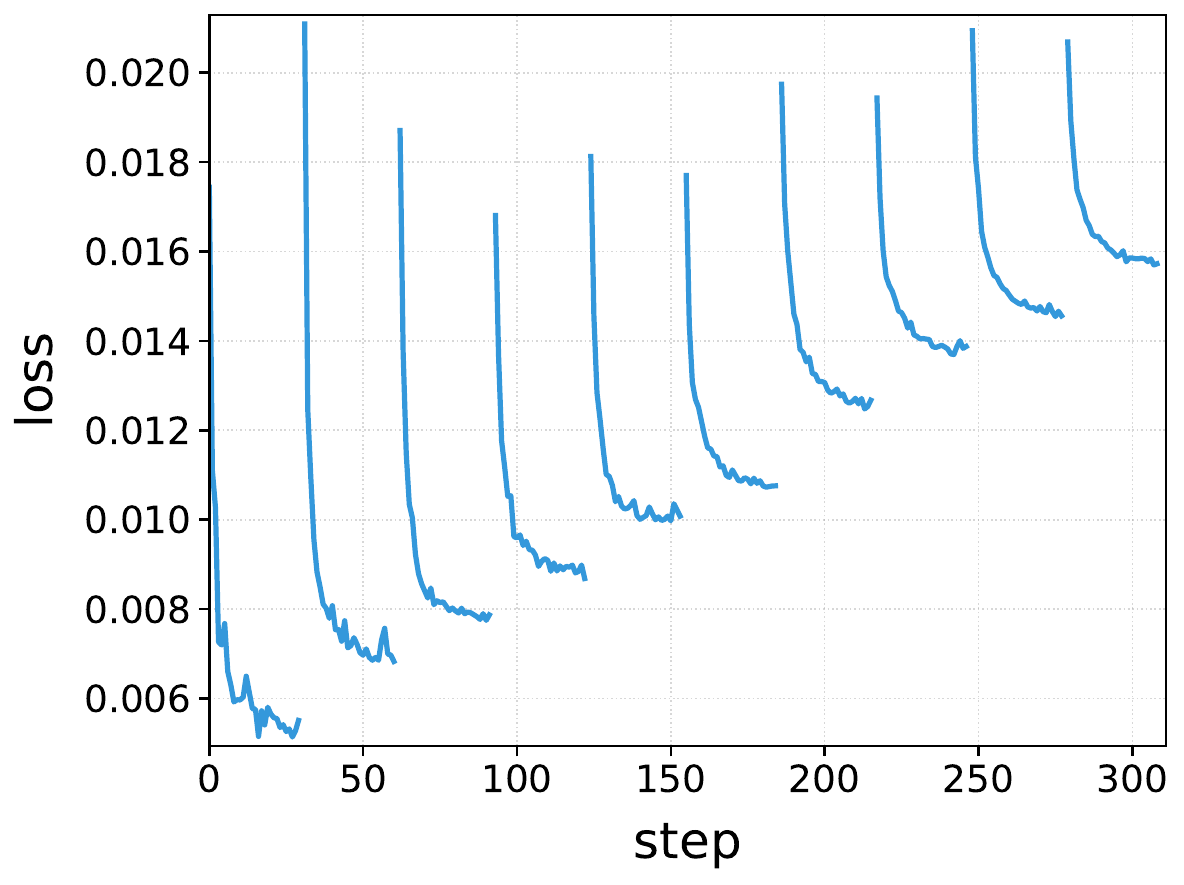}
    \end{minipage}
    \hfill
    \begin{minipage}[b]{0.24\textwidth}
        \includegraphics[width=\textwidth]{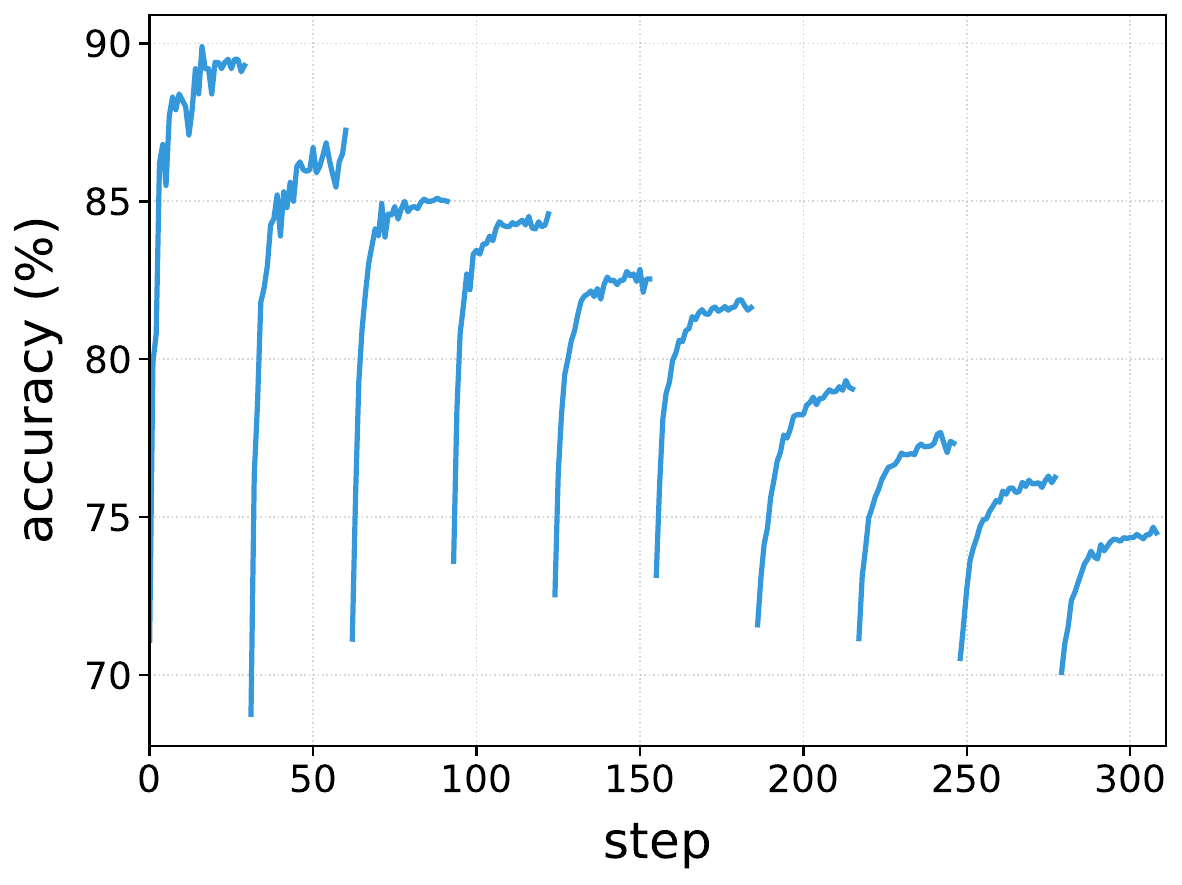}
    \end{minipage}
    \caption{Training curves for $T=10$ on Tiny-ImageNet when $\alpha=12$.}
    \label{fig:curve_T10}
\end{figure}

\begin{figure}[!ht]
    \centering
    \begin{minipage}[b]{0.24\textwidth}
        \includegraphics[width=\textwidth]{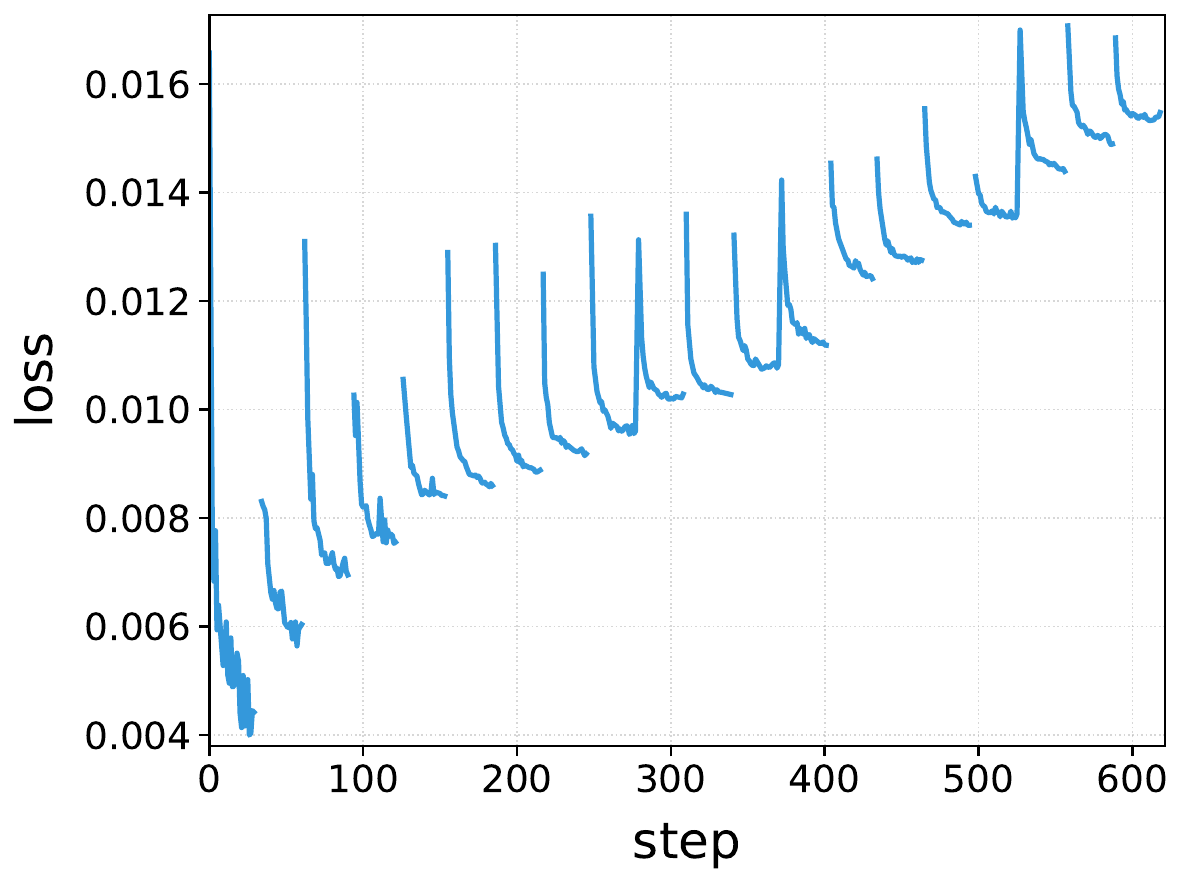}
    \end{minipage}
    \hfill
    \begin{minipage}[b]{0.24\textwidth}
        \includegraphics[width=\textwidth]{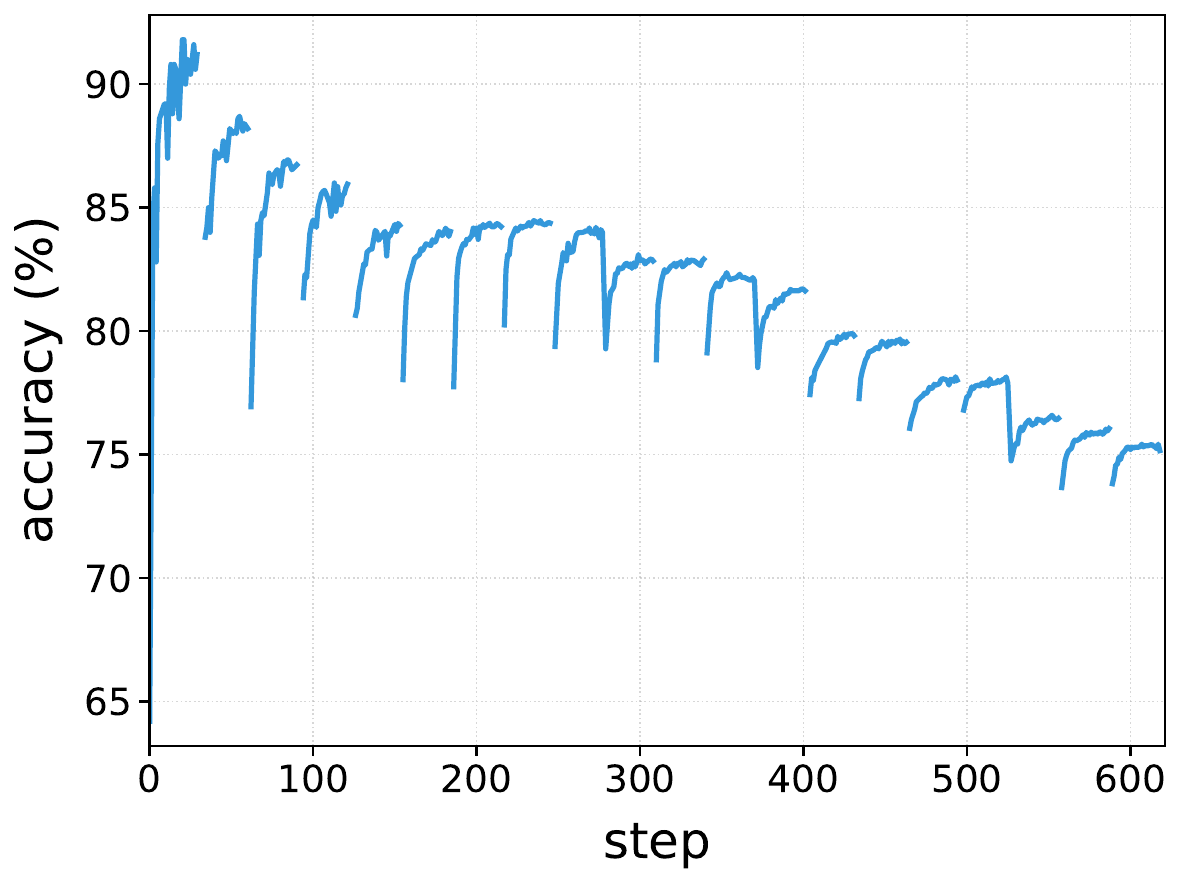}
    \end{minipage}
    \caption{Training curves for $T=20$ on Tiny-ImageNet when $\alpha=12$.}
    \label{fig:curve_T20}
\end{figure}

\begin{figure}[!ht]
    \centering
    \begin{minipage}[b]{0.24\textwidth}
        \includegraphics[width=\textwidth]{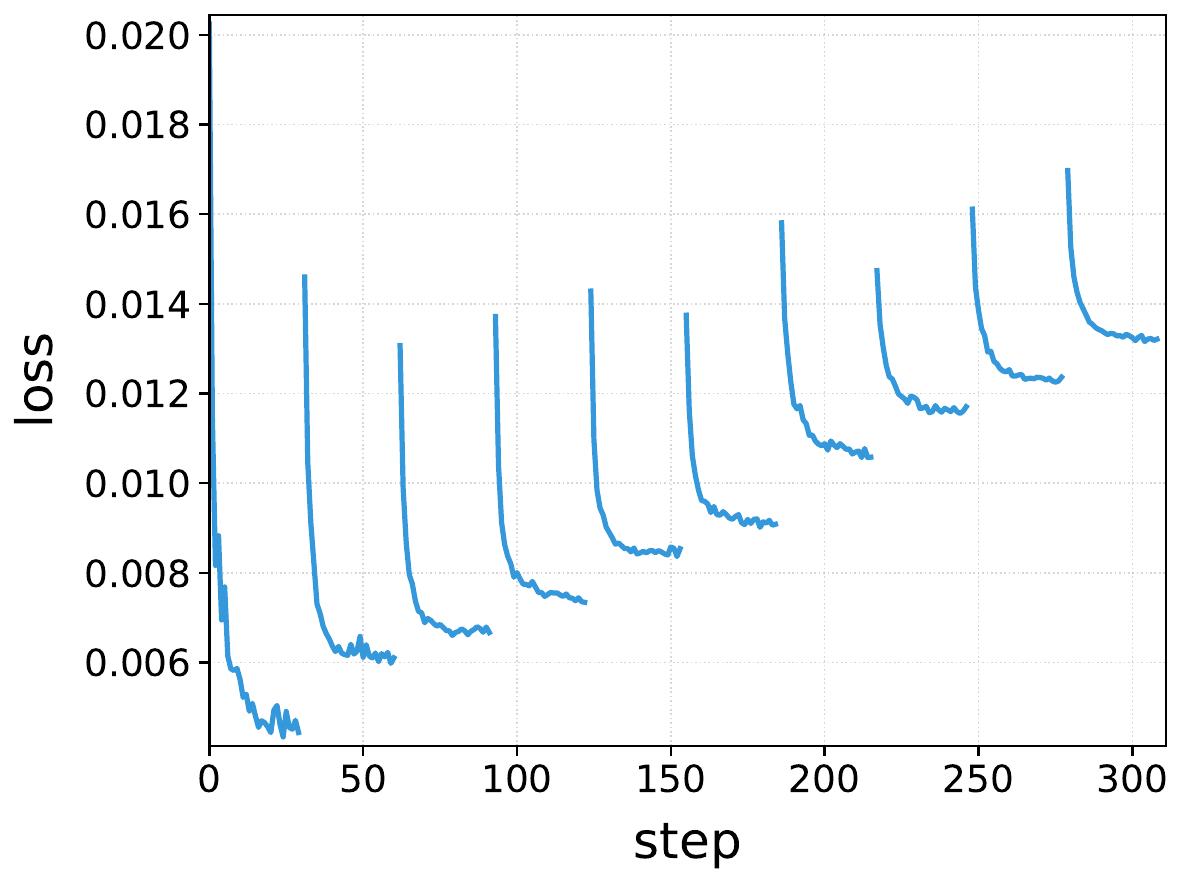}
    \end{minipage}
    \hfill
    \begin{minipage}[b]{0.24\textwidth}
        \includegraphics[width=\textwidth]{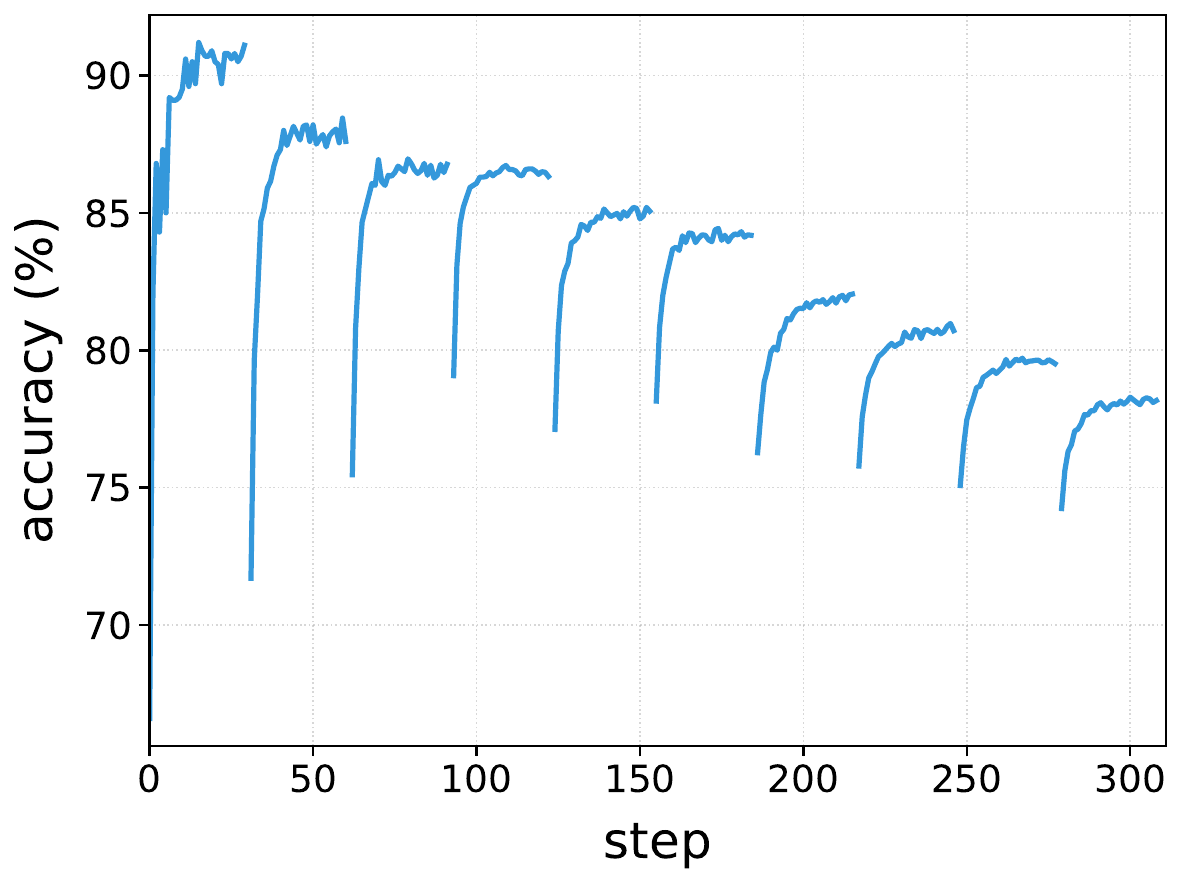}
    \end{minipage}
    \caption{Training curves for $T=10$ on Tiny-ImageNet when $\beta=0.5$.}
    \label{fig:curve_T10_beta05}
\end{figure}

\textbf{Training Curves}.
To illustrate the convergence of proposed Fed-TaLoRA. 
we plot some selected training curves for $T=5$ (Figure \ref{fig:curve_T5}), $T=10$ (Figure \ref{fig:curve_T10}) and $T=20$ (Figure \ref{fig:curve_T20}) on Tiny-ImageNet when $\alpha=12$, and $T=10$ on Tiny-ImageNet when $\beta=0.5$ (Figure \ref{fig:curve_T10_beta05}).
We can see that proposed Fed-TaLoRA can converge at different tasks' lengths representing the complexity of downstream tasks regardless of distribution-based label imbalance and quantity-based label imbalance.


\end{document}